\documentclass[10pt,twocolumn,letterpaper]{article}

\usepackage{cvpr}              
\usepackage{dsfont}
\usepackage{graphicx} 
\usepackage{float}
\usepackage[dvipsnames]{xcolor}

\definecolor{cvprblue}{rgb}{0.21,0.49,0.74}
\newcommand{\ptitle}[1]{\noindent\textbf{#1}\hspace{5pt}}

\usepackage[pagebackref,breaklinks,colorlinks,allcolors=cvprblue]{hyperref}

\def\paperID{3144} 
\def\confName{CVPR}
\def\confYear{2025}

\title{Progressive Correspondence Regenerator for Robust 3D Registration}

\author{Guiyu Zhao\textsuperscript{1$\ast$}, 
    Sheng Ao\textsuperscript{2}\thanks{Equal contribution} ,
    Ye Zhang\textsuperscript{2},
    Kai Xu\textsuperscript{3}
    Yulan Guo\textsuperscript{2$\dagger$}
    \\
    \textsuperscript{1}Beijing Institute of Technology 
    \textsuperscript{2}Sun Yat-sen University 
    \textsuperscript{3}National University of Defense Technology
}

\begin{document}
\maketitle

\begin{abstract}

    Obtaining enough high-quality correspondences 
    is crucial for robust registration. Existing correspondence refinement methods mostly follow the paradigm of outlier removal, which either fails to correctly identify the accurate correspondences under extreme outlier ratios, or select too few correct correspondences to support robust registration. To address this challenge, we propose a novel approach named Regor, which is a progressive correspondence regenerator that generates higher-quality matches whist sufficiently robust for numerous outliers. In each iteration, we first apply prior-guided local grouping and generalized mutual matching to generate the local region correspondences. A powerful center-aware three-point consistency is then presented to achieve local correspondence correction, instead of removal. Further, we employ global correspondence refinement to obtain accurate correspondences from a global perspective. Through progressive iterations, this process yields a large number of high-quality correspondences. Extensive experiments on both indoor and outdoor datasets demonstrate that the proposed Regor significantly outperforms existing outlier removal techniques. More critically, our approach obtain 10 times more correct correspondences than outlier removal methods. As a result, our method is able to achieve robust registration even with weak features. The code will be released.
\end{abstract}

\section{Introduction}



\begin{figure}[ht]
    \centering{\includegraphics[width=1.0\linewidth]{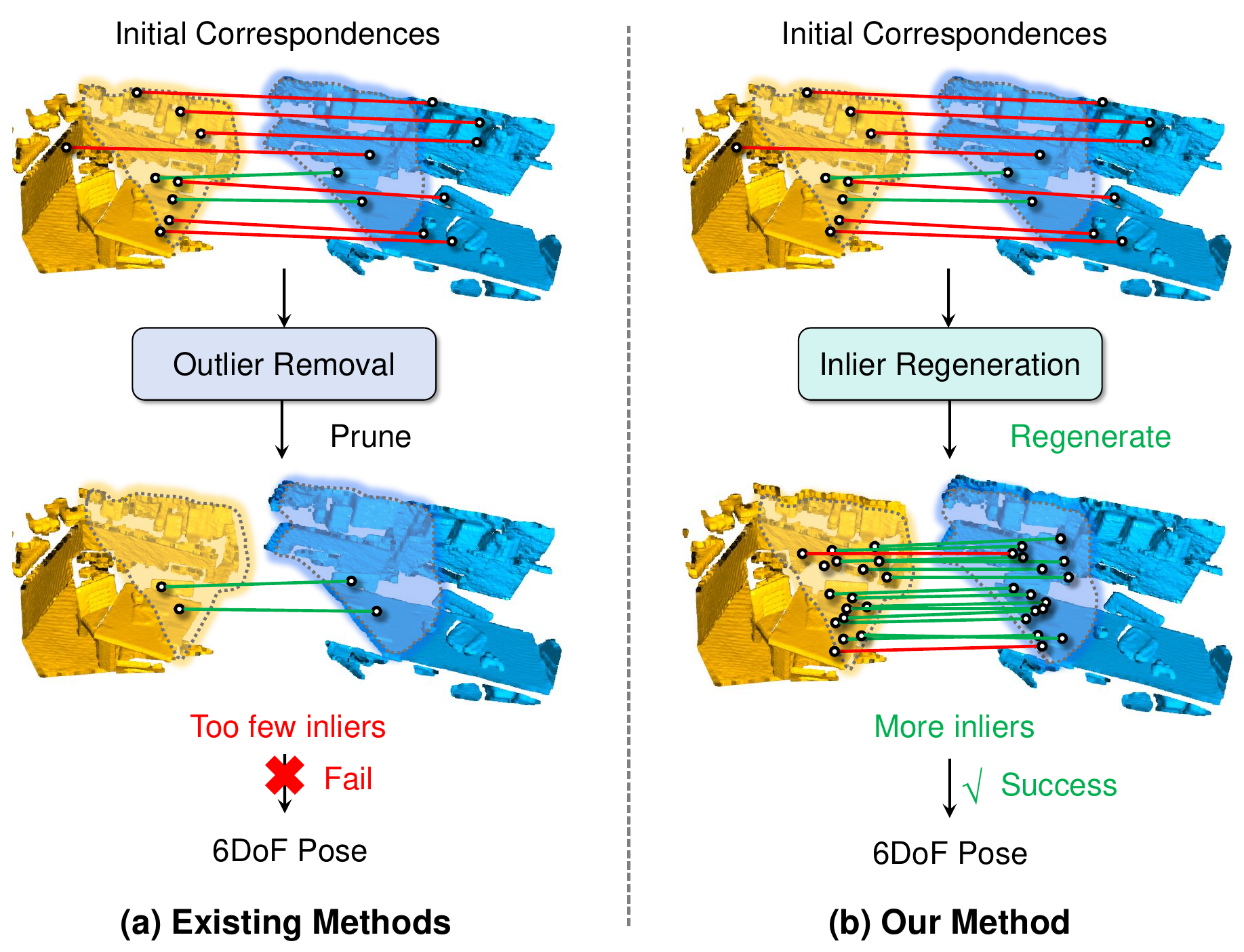}}  
    \caption{
        Difference between our method and existing outlier removal techniques. Outlier removal methods perform top-down pruning, producing an optimized set of correspondences that is a subset of the initial correspondences. When the initial inliers are few, this approach can obtain only a limited number of inliers, which is insufficient for robust pose estimation. In contrast, our method takes a bottom-up approach, using a regeneration strategy to generate more inliers, thereby effectively addressing this issue.
        }
    \label{process}
    \vspace{-5pt}
\end{figure}

Point cloud registration~\cite{huang2021comprehensive} is a fundamental and critical task in 3D computer vision, with widespread applications across various domains such as robotic localization~\cite{xiong2023speal, yin2024survey}, object pose estimation~\cite{yu2024learning}, and remote sensing~\cite{huang2021comprehensive}. Given two point clouds scanned from different perspectives, 
point cloud registration aims to estimate a rigid transformation to align these two point clouds. 
With the advancement of deep learning and feature metric learning, feature-based point cloud registration 
methods~\cite{ao2021spinnet,huang2021predator,yu2021cofinet,qin2022geometric} have garnered significant attention and achieved promising results. However, in scenarios characterized by low overlap, high noise, and strong self-similarity, direct feature matching often leads 
to a large number of outliers, which poses a great challenge for point cloud registration.

Currently, the mainstream approach to address the issue of numerous outliers in point cloud registration 
is to filter wrong correspondences. Existing outlier removal methods~\cite{chen2022sc2,PointDSC,lu2021hregnet,zhang20233d,jiang2023robust} can 
be categorized into two types: geometry-based methods and learning-based methods. 
Geometry-based methods~\cite{fischler1981random,olsson2008branch,zhang20233d,chen2022sc2}
typically remove outliers according to whether point correspondences satisfy geometric consistency. 
Learning-based methods~\cite{PointDSC,lu2021hregnet,jiang2023robust} treat this task as a classification problem to distinguish between outliers and inliers. 
However, in challenging scenarios, these correspondence filtering methods all suffer from a common problem: when initial correspondences only contains few inliers, estimating the correct rigid transformation between two point clouds is extremely difficult even if all inliers are identified.

In this paper, we aim to design a new 3D registration architecture, which is robust to few inliers and is able to estimate precise 6DoF pose.
As shown in Figure~\ref{process},
different from existing outlier removal methods~\cite{chen2022sc2,PointDSC,zhang20233d,chen2023sc2}
that focus on pruning correspondences, our approach clearly regenerate more high-quality correspondences to achieve robust registration. The key insight is to progressively generate correspondences and leverage the positional priors of correspondences from the previous iteration to guide more refined local regeneration, thereby obtaining more high-quality point correspondences.

Our method, named Regor, follows a progressive iterative framework composed of three main modules: Local Grouping and Rematching, Local Correspondence Refinement and Global Correspondence Refinement. In each iteration, \textbf{Local Grouping and Rematching} leverages \emph{prior-guided local grouping} to gradually reduce the matching space and \emph{generalized mutual matching} to perform reliable local matching. 
The \textbf{Local Correspondence Refinement} introduces an effective \emph{center-aware three-point consistency} to mine inliers and update local correspondences. The \textbf{Global Correspondence Refinement} is employed to perform further optimization from a global perspective. Note that, each iteration gradually increases the number of inliers, instead of only removing outliers. 
Our method achieves state-of-the-art performance on both the 3DMatch~\cite{zeng20173dmatch} and KITTI~\cite{KITTIdataset} datasets. Notably, the proposed Regor demonstrates a strong ability to generate correspondences, yielding up to 10 times more correct matches than existing outlier removal methods~\cite{chen2022sc2,zhang20233d,chen2023sc2}. 
The main contributions of this paper are as follows:
\begin{itemize}
    \setlength{\itemsep}{0pt}
    \setlength{\parsep}{0pt}
    \setlength{\parskip}{0pt}
    \item[$\bullet$] 
    We propose a novel registration approach named Regor to obtain massive high-quality point correspondences, overcoming the challenge of scarce inliers by progressively generating correspondences.

    \item[$\bullet$] 
    Our method achieves state-of-the-art performance. In particular, the number of correct correspondences is 10 times higher than that of outlier removal methods.

    \item[$\bullet$]
    Benefiting from the proposed regenerator, our approach achieves robust registration even with weak features.
    
 \end{itemize}

\section{Related Work}

\subsection{3D Feature Matching}
Feature matching is a crucial step in 
correspondence-based point cloud registration~\cite{FPFH,zeng20173dmatch,ao2021spinnet,huang2021predator,yu2021cofinet,qin2022geometric,yu2023peal}. 
According to the approach of feature extraction, feature matching methods 
can be categorized into two types: traditional descriptor-based methods~\cite{PFH, FPFH, salti2014shot} and 
learning-based methods~\cite{ao2021spinnet,huang2021predator,yu2021cofinet,qin2022geometric,yu2023peal}. 
Before the widespread adoption of deep learning, features were manually designed descriptors tailored to represent local information. 
They are classified into LRF-based methods~\cite{guo2013rotational, salti2014shot} and LRF-free methods~\cite{PFH, FPFH} 
based on whether a local reference frame (LRF) is needed. With the advancements in deep learning, 
numerous learning-based methods have emerged. Based on the representation strategy, 
they can be further divided into patch-based methods~\cite{gojcic2019perfect,ao2021spinnet,zhao2023spherenet} 
and fragment-based methods~\cite{choy2019fully,huang2021predator,yu2021cofinet,gath1989unsupervised}. 
SpinNet~\cite{ao2021spinnet} and BUFFER~\cite{ao2023buffer} have achieved excellent generalization ability through 3D cylindrical convolution. Additionally, CoFiNet~\cite{yu2021cofinet} and GeoTransformer~\cite{qin2022geometric} propose a coarse-to-fine matching strategy and a Transformer-based framework, which has garnered widespread attention. Subsequent methods~\cite{yang2022one, yu2023rotation, yu2023peal, jin2024multiway} have further 
improved performance by incorporating position encoding and prior information into this framework.
ODIN~\cite{jin2024multiway} achieves the best performance through diffusion strategy and global optimization.

\subsection{Geometry-based Outlier Removal}
Geometry-based outlier removal involves fitting models from noisy correspondences using the geometric properties of 3D scenes to estimate poses. The most classic method is random sampling consensus (RANSAC)~\cite{fischler1981random}, which 
samples correspondences multiple times and validates to 
obtain the best pose, thus mitigating the impact of noisy correspondences. Subsequently, numerous variants of RANSAC~\cite{barath2018graph, barath2022space, chum2008optimal, schnabel2007efficient} have been proposed to address issues 
of time consumption and instability. TEASER~\cite{yang2020teaser} introduces the truncated least squares (TLS) cost to solve for pose and employ rotation invariant measurements to handle outliers. Then, SC$^2$-PCR~\cite{chen2022sc2, chen2023sc2} propose second-order consistency for robust outlier removal. In recent years, methods~\cite{zhang20233d, yang2023mutual} based on 
geometric consistency have been widely applied in point cloud registration, leveraging graph-theoretic approaches~\cite{eppstein2010listing} and geometric consistency to 
remove outliers, thereby achieving robust registration.

\begin{figure*}[t]
    \centering{\includegraphics[width=1.0\textwidth]{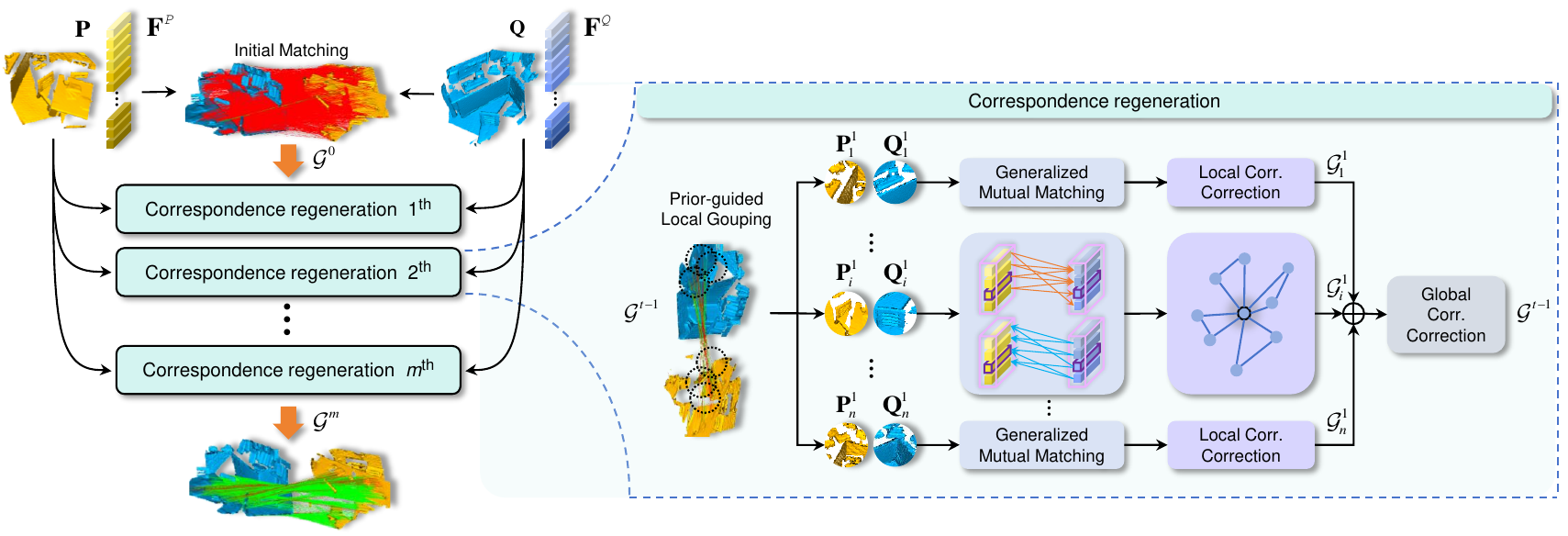}}  
    \caption{
       Overall framework of our method.
       We extract features from the original point cloud, 
       obtaining features $\mathbf{{F} }^{\mathcal{P}}$ and $\mathbf{{F} }^{\mathcal{Q}}$ 
       as input for our method. 
       Subsequently, a progressive 
       process is applied to iteratively regenerate more accurate and denser 
       correspondences $\mathcal{{G}}^{t}$. At each iteration, the output correspondences $\mathcal{{G}}^{t-1}$ from the previous 
       iteration serve as input. Firstly, prior-guided local grouping is employed to sample 
       seed corresponding points and form local correspondence regions 
       $\mathbf{P}^{t}_i$ and $\mathbf{Q}^{t}_i$. Then, for each pair of local correspondence regions, 
       generalized mutual matching is performed to get new correspondences. Next, these correspondences are refined locally and globally using our center-aware three-point consistency, 
       followed by a merging operation $\oplus$ of local correspondences $\mathcal{{G}}^{t}_{i}$ using a hash table.
       Finally, using these refined correspondences,
       we achieve robust and accurate transformation estimation $\mathbf T\{\mathbf R,\mathbf t\}$ only with SVD.
       }
    \label{fig1}
 \end{figure*}

\subsection{Learning-based Outlier Removal}
Recent researches~\cite{3DRegNet, PointDSC, jiang2023robust} have integrated 
deep learning into outlier removal and pose estimation, 
showcasing promising performance through training. 
Inspired by image matching~\cite{CN-Net}, 3DRegNet~\cite{3DRegNet} adapts CN-Net~\cite{CN-Net} to remove outliers in point cloud registration. Subsequently, PointDSC~\cite{PointDSC} embeds the spatial consistency 
into feature maps and employs neural spectral matching to estimate confidence of correspondences, 
achieving robust registration. VBReg~\cite{jiang2023robust} utilizes variational Bayesian inference 
to achieve better outlier removal. Hunter~\cite{yao2023hunter} addresses severe outlier issues 
by introducing higher-order consistency using hypergraphs. However, while both geometry-based 
and learning-based methods have shown effectiveness in outlier removal, 
they inherently rely on the initial correspondences and only trim down on them, 
which poses challenges when initial inliers are scarce.

\section{Method}

\subsection{Problem Formulation}
Point cloud registration is aligning two point clouds $\mathbf{P}$ and $\mathbf{Q}$, captured from different perspectives, by estimating a pose transformation $\mathbf T\{\mathbf R,\mathbf t\}$ where $\mathbf R$ is a rotation matrix and $\mathbf t$ is a translation vector. Feature-based point cloud registration estimates correspondences through feature extraction and matching, then minimizes the average Euclidean distance between these correspondences, as shown in Eq.~\ref{eq1}:
\begin{equation}
    \underset{\mathbf{R}\in SO(3), \mathbf{t}\in \mathbb{R}^3}{\arg \min } \sum_{\left({\mathbf{p}}_{x_i}, {\mathbf{q}}_{y_i}\right) \in \mathcal{I}} \left\|\mathbf{R} \cdot {\mathbf{p}}_{x_i}+\mathbf{t}-{\mathbf{q}}_{y_i}\right\|_2^2.
\label{eq1}
\end{equation}
In this study, unlike outlier removal methods that perform ``subtraction'' filtering on initial correspondences, 
resulting in a subset of initial correspondences highly dependent on them, 
our aim is to generate more higher-quality correspondences $\mathcal{G}^{t}$  where $t=1, 2, ..., m$ represents the current iterative stage. We define the regeneration process as $\mathcal{G}^{t} = \varTheta (\mathcal{G}^{t-1};\mathbf{P}, \mathbf{Q}, \varphi^{t} )$
where $\varTheta$ is the operation of the correspondence regeneration, and $\varphi^{t}$ represents the parameter used in the regeneration of the $t$th stage.

\subsection{Progressive Regeneration: Key Insight}
Existing methods focus on pruning the initial correspondences to remove outliers and retain inliers. 
Due to common problems such as weak feature representation and low overlap, the proportion of correct initial correspondences is typically very small. Even all correct correspondences are identified, achieving accurate registration remains challenging. To this end, we design a novel progressive correspondence regeneration strategy to address the scarcity of initial inliers and the frequent failures of correspondence filtering under extreme outlier ratios.

The key idea of our method is to progressively regenerate better correspondences in a radius-variable local sphere.
We use the positional priors of correspondences generated in the previous stage to guide the creation of the local sphere.
In each local sphere, we regenerate more correspondences with higher quality.
Compared to direct global matching, re-matching and refining within these local 
spheres significantly reduce the problem's scale and search space, resulting in more precise matches and 
reduced computational time. As the quality of the correspondence gets higher, we progressively reduce the size of the local spheres to 
achieve exact convergence of the correspondences. This approach not only generates a large number of new dense correspondences but also incrementally enhances their accuracy.

\subsection{Local Grouping and Rematching}\label{Grouping}
In this module, we implement the core of each iteration: local correspondence generation. First, we introduce prior-guided local grouping to leverage priors from the previous iteration. Then, we propose a generalized mutual matching technique to ensure the generation of a sufficient number of inliers even under high outlier rates.

\ptitle{Prior-guided Local Grouping.}
The input to our method is the correspondence $\mathcal{G}^{t-1}$ of the previous stage.
Efficiency is crucial for our method due to the need for iterative processing. 
Therefore, we sample $\mathcal{G}^{t-1}$ 
using efficient random sampling~\cite{hu2020randla} 
to derive seed correspondences $ \mathcal{\widetilde{G}}^{t-1}$. 
We then utilize 
the positional priors of each seed correspondence to guide the construction of local regions.  
Specifically, we conduct a radius $ r^{t}$ nearest neighbor search centered on each seed correspondence $ {\widetilde{g}}^{t-1}_i \in \mathcal{\widetilde{G}}^{t-1}$
to obtain local point clouds  $\mathbf{P}^{t}_i$ and $\mathbf{Q}^{t}_i$, $i=1, 2, ..., n$:
\begin{equation}
    \mathbf{P}^{t}_i = \mathrm{RNN}(\mathbf{p}^{t-1}_i; r^{t}, \mathcal{\widetilde{G}}^{t-1}),   \mathbf{Q}^{t}_i = \mathrm{RNN}(\mathbf{q}^{t-1}_i; r^{t}, \mathcal{\widetilde{G}}^{t-1}),
\end{equation}
where $\mathrm{RNN}(\mathbf{p}; r, \mathcal{{G}})$ is a radius nearest neighbor search on $\mathcal{{G}}$ with $\mathbf{p}$ as the center and $r$ as the radius. 
$\mathbf{p}^{t-1}_i$ and $\mathbf{q}^{t-1}_i$ satisfy $ {\widetilde{g}}^{t-1}_i =(\mathbf{p}^{t-1}_i, \mathbf{q}^{t-1}_i)$.
We perform the above grouping operation for each pair of seed corresponding points and obtain $n$ pairs of local region
$(\mathbf{P}^{t}_i, \mathbf{Q}^{t}_i)$.

\begin{figure}[tb]
    \centering{\includegraphics[width=1.0\linewidth]{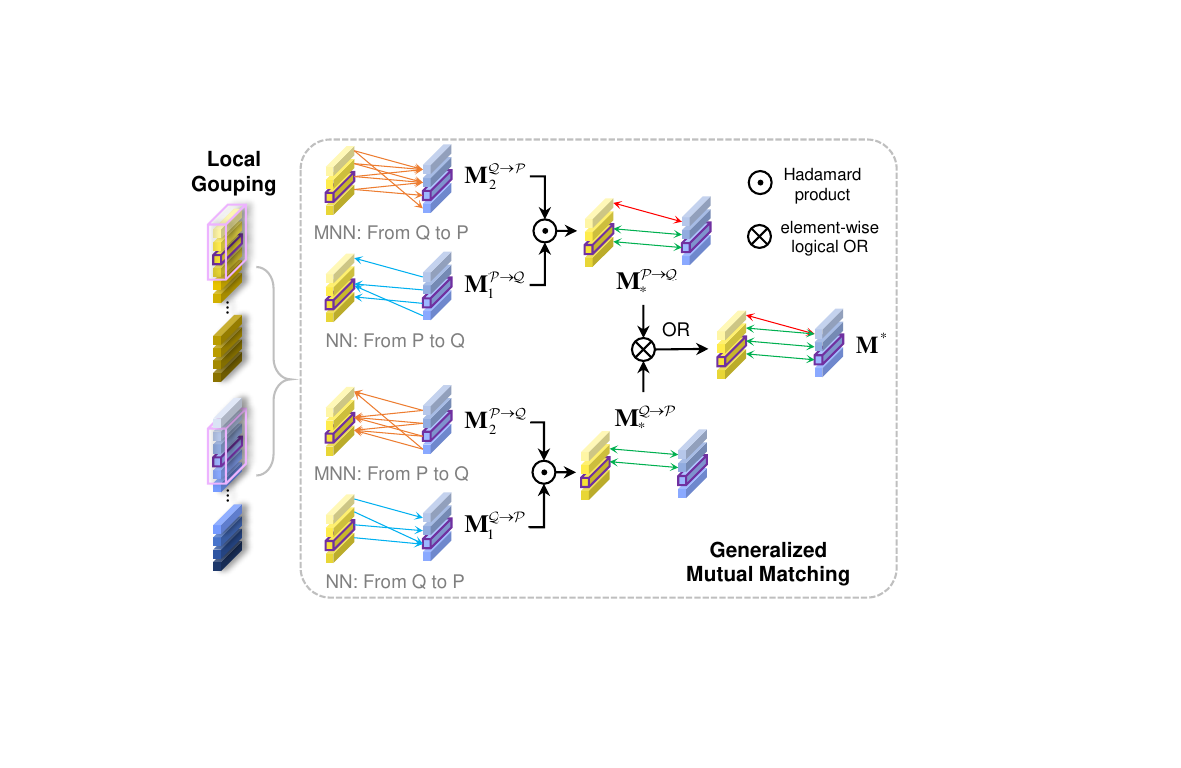}}  
    \caption{
        Illustration of GMM.
        The orange, blue, green, and red lines are the MNN correspondences, NN correspondences, correct correspondences, and wrong correspondences, respectively.
       }
    \label{GMM}
\end{figure}

\ptitle{Generalized Mutual Matching.}
The next step is to generate new point correspondences from each pair of local region by feature matching. 
However, the features in small local regions are usually similar, which results in traditional nearest-neighbor (NN) matching~\cite{PointSignatures} producing a large number of false correspondences. Unfortunately, mutual NN matching~\cite{spinimage,3D-3D} may even fail to find any correspondence at all. To address this issue, we propose a generalized mutual matching to relax the strict mutual constraint, thereby improving the robustness of the local correspondences.


Given point cloud $\mathbf{Q}^{t}_i$ as the query, the NN matching matrix $\mathbf{M}^{\mathcal{P}\rightarrow\mathcal{Q} }_1$ and multi-nearest-neighbor (MNN) matching matrix $\mathbf{M}^{\mathcal{P}\rightarrow\mathcal{Q} }_2$ are obtained by performing a k-nearest-neighbor (KNN) search in feature space, respectively: 
\begin{equation}
    \begin{aligned}
    &\mathbf{M}^{\mathcal{P}\rightarrow\mathcal{Q} }_1(m,n)=\operatorname{NN}(\mathbf{F}(\mathbf{p}_m), \mathbf{F}(\mathbf{q}_n)), \\ 
    &\mathbf{M}^{\mathcal{P}\rightarrow\mathcal{Q} }_2(m,n)=\operatorname{MNN}(\mathbf{F}(\mathbf{p}_m), \mathbf{F}(\mathbf{q}_n)),
    \end{aligned}
\end{equation}
where $\forall \mathbf{p}_m \in \mathbf{P}^{t}_i, \mathbf{q}_n \in \mathbf{Q}^{t}_i$ and $\mathbf{F}(\cdot)$ represents the mapping from points to features. $\operatorname{NN}(\cdot)$ and $\operatorname{MNN}(\cdot)$ represent the indicator functions for NN and MNN matching, respectively.  If the matching condition is satisfied, the value is set to 1. 
Then, we conduct mutual matching from 
$\mathbf{P}^{t}_i$ to $\mathbf{Q}^{t}_i$
and from $\mathbf{Q}^{t}_i$ to $\mathbf{P}^{t}_i$, resulting in four matching matrices $\mathbf{M}^{\mathcal{P}\rightarrow\mathcal{Q} }_1$, $\mathbf{M}^{\mathcal{P}\rightarrow\mathcal{Q} }_2$, $\mathbf{M}^{\mathcal{Q}\rightarrow\mathcal{P} }_1$, and $\mathbf{M}^{\mathcal{Q}\rightarrow\mathcal{P} }_2$.

By computing the Hadamard product between the NN and MNN matching matrices, we obtain the mutual matching matrices  $\mathbf{M}^{\mathcal{P}\rightarrow\mathcal{Q} }_*$, $\mathbf{M}^{\mathcal{Q}\rightarrow\mathcal{P} }_*$. Further, we perform an element-wise logical OR operation to relax the constraint and get the generalized mutual matching matrix:
\begin{equation}
    \mathbf{M}^{*} = \left( \mathbf{M}^{\mathcal{P}\rightarrow\mathcal{Q} }_1 \odot \mathbf{M}^{\mathcal{Q}\rightarrow\mathcal{P} }_2  \right) \otimes \left( \mathbf{M}^{\mathcal{Q}\rightarrow\mathcal{P} }_1 \odot \mathbf{M}^{\mathcal{P}\rightarrow\mathcal{Q} }_2  \right),
\end{equation}
where operation $\odot$ and $\otimes$ denote Hadamard product and element-wise logical OR, respectively.
It is important to note that, to avoid a large number of outliers, we discard the strategy of directly applying mutual matching only based on MNN. Finally, using the matching matrix $\mathbf{M}^{*}$, we establish $n$ sets of local correspondences $\mathcal{{G}}^{t}_i$.


\subsection{Local Correspondence Refinement}

Because some corresponding local regions may not overlap exactly, the rematching in local regions still introduces some erroneous correspondences. Therefore, we further enhance the quality 
of the regenerated correspondences through local correspondence refinement.


\begin{figure}[ht]
    \centering{\includegraphics[width=1.0\linewidth]{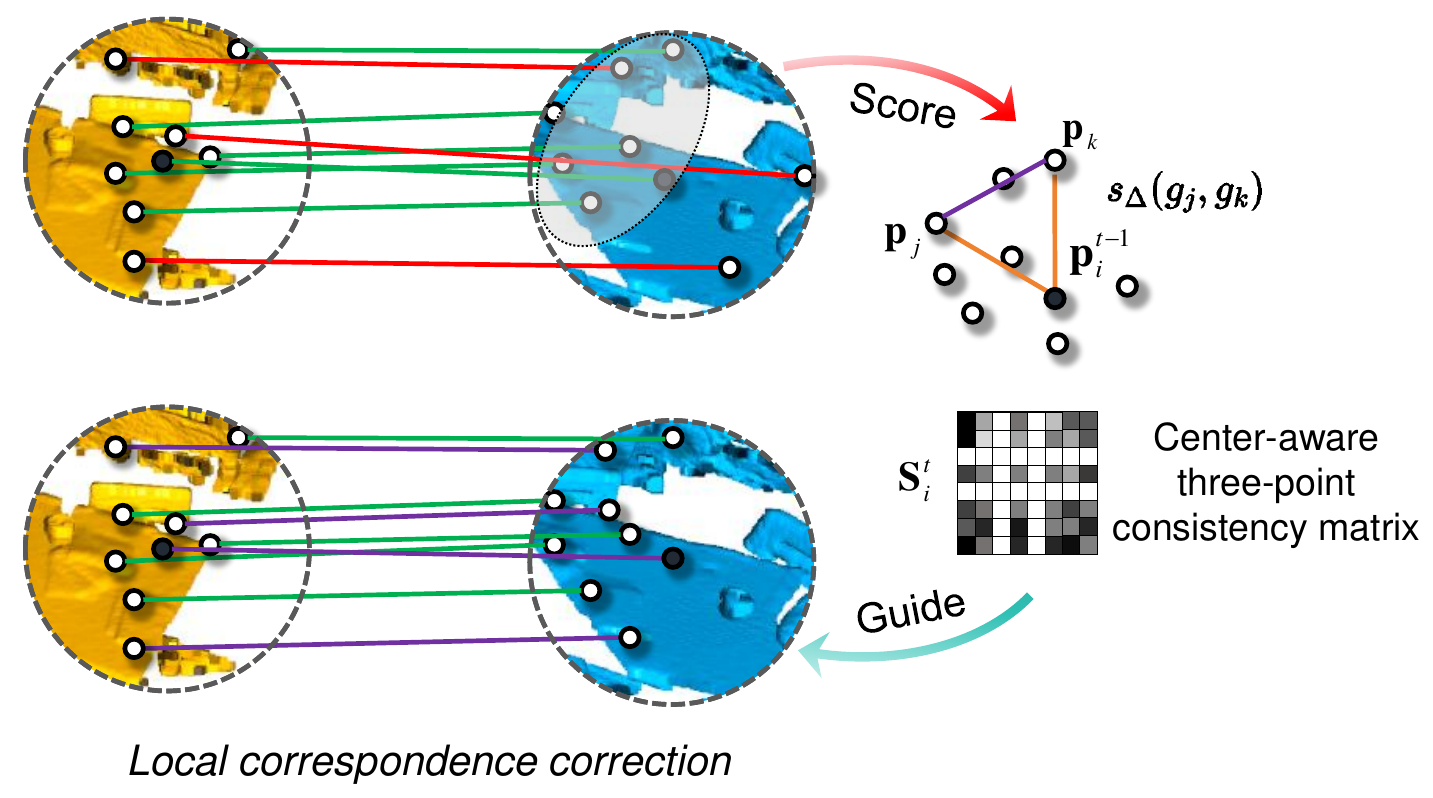}}  
    \caption{
        Illustration of the local correspondence correction.
        Yellow and blue areas represent a pair of local point clouds. 
        The purple line indicates the correct correspondence after correction.
       }
    \label{LGCR}
\end{figure}

\ptitle{Center-aware Three-point Consistency.}
Note that, our seed correspondences $g^{t\!-\!1}_i\!=\!(\mathbf{p}^{t\!-\!1}_i, \mathbf{q}^{t\!-\!1}_i)$ (central points) are derived from the previous stage and exhibit high accuracy. Therefore, they can serve as prior information to guide the optimization of local correspondence refinement.
Based on this, we propose an efficient center-aware three-point consistency (CTC) to search inliers. 

We first utilize efficient translation and rotation invariants~\cite{yang2020teaser} to describe the consistency $s_{\sigma}(g_i, g_j)$ between two point pairs 
$g_i=(\mathbf{p}_i, \mathbf{q}_i)$ and $g_j=(\mathbf{p}_j, \mathbf{q}_j)$: 
\begin{equation}
    s_{\sigma}(g_i, g_j)= 
    \mathds {1} \left(\Big| ||\mathbf{p}_i-\mathbf{p}_j||_2 
    -||\mathbf{q}_i-\mathbf{q}_j||_2 \Big| \leqslant {\sigma}
    \right).
\end{equation}
Next, we leverage the prior information of accurate center points to compute the center-aware consistency between $g_j$ and $g_k$, denoted as $s_{\sigma}(g_j, g^{t-1}_i) \cdot s_{\sigma}(g^{t-1}_i, g_k)$. Additionally, we account for the possibility that seed correspondences may not always be correct. Relying solely on the aforementioned criterion could allow errors in the seed correspondences to negatively impact the optimization of the entire region. To mitigate this, we introduce a strict point-pair constraint $s_{\frac{\sigma}{2}}(g_j, g_k)$.
CTC is computed as:
\begin{equation}
    s_{\varDelta}(g_j, g_k)
    = \left(s_{\sigma}(g_j, g^{t-1}_i) \cdot s_{\sigma}(g^{t-1}_i, g_k)\right)||s_{\frac{\sigma}{2}}(g_j, g_k),
\end{equation}
where $||$ represents operator OR and it
means if either condition is satisfied, the correspondence is considered correct.
Therefore, we get the score matrix $\mathbf{S}^t_{i}=[s_{\varDelta}(g_j, g_k)]_{jk} \in \mathbb{R}^{N \times N}$ of the local correspondences $\mathcal{{G}}^{t}_i \in \mathbb{R}^{N \times 2}$.
To evaluate the overall quality of the local region,
we define a local overall score:
\begin{equation}
    Score(\mathcal{G}^{t}_i)
    = \frac{|| \mathbf{S}^t_{i}||_1}{a\cdot N},
\end{equation}
where $||\cdot||_1$represents the $L_1$-norm of a matrix and $a \in (0, 1)$ denotes the threshold for the proportion of correct correspondences. 
Essentially, we consider regions where 
the proportion of correct correspondences exceeds $a$ as correct correspondence regions. 
This is Theorem~\textcolor{red}{1}, and then we give its proof in the Appendix.

\ptitle{Local Correspondence Correction.}
Within each correct corresponding region, we identify a set of precise correspondences $\mathcal{\widehat {G}}^{t}_i$ to 
guide the refinement of correspondences across the entire local region.  Specifically, based on the CTC score, we identify precise correspondences $\mathcal{\widehat {G}}^{t}_i$ with high scores.  
Subsequently, using these correspondences, we estimate the local transformation  $\mathbf{T}_i^{t}\{\mathbf{R}_i^{t}, \mathbf{t}_i^{t}\}$:
\begin{equation}
    \mathbf{R}_i^{t}, \mathbf{t}_i^{t}=\min_{\mathbf{R}, \mathbf{t}} \sum\nolimits_{\left({\mathbf{p}}_{j}, {\mathbf{q}}_{j}\right) \in \widehat{\mathcal{ {G}}}_i^{t}} \left\|\mathbf{R} \cdot {\mathbf{p}}_{j}+\mathbf{t}-{\mathbf{q}}_{j}\right\|_2^2.
\end{equation}
Leveraging this local pose $\mathbf{T}_i^{t}\{\mathbf{R}_i^{t}, \mathbf{t}_i^{t}\}$, we search 
for the nearest points $\mathbf{q}^{\prime}$ in the target local point cloud $\mathbf{Q}^{t}_i$ for each point $\mathbf{p}$ in the source local point cloud $\mathbf{P}^{t}_i$:
\begin{equation}
\begin{aligned}
&\mathbf{q}^{\prime}= \mathop{\arg\min}\limits_{\mathbf{q}_j \in \mathbf{Q}^{t}_i} ||\mathbf{R}_i^{t}\mathbf{p}+\mathbf{t}_i^{t}-\mathbf{q}_j||_2, \\
        s.t. &||\mathbf{R}_i^{t}\mathbf{p}+\mathbf{t}_i^{t}-\mathbf{q}^{\prime}||_2\leqslant {\sigma_d}, \forall(\mathbf{p},\mathbf{q}) \in \mathcal{{G}}^{t}_i.
\end{aligned}
\end{equation}
Through the one-way refinement of $\mathbf{Q}$, we update our local correspondences $\mathcal{{G}}^{t}_i \leftarrow \Omega_l (\mathcal{{G}}^{t}_i)=\{(\mathbf{p}, \mathbf{q}^{\prime})\} $
where $\Omega_l$ represents the operation of local correspondence correction.

\ptitle{Theorem 1.} \label{Theorem1}
    \emph{
        Assuming event $\Phi $, where the consistency of correct correspondences is greater than the maximum consistency of incorrect correspondences, 
        has a probability ${P}(\Phi)\approx  1 $,
        if the score of the local correspondences satisfies $Score(\mathcal{G}^{t}_i)\geqslant  1$, then the proportion of correct correspondences $b \geqslant a$.
        Theorem 1 is proved and explained in detail in the appendix.
    }

\subsection{Global Correspondence Refinement}
Through local refinement, we obtain $n$ sets of refined local correspondences. However, consistency across sets is not fully aligned. Therefore, we need a global merging and refinement step to complete the global correspondence optimization.
Since there is no seed correspondence at the global scale, we use second-order consistency~\cite{chen2022sc2}.

\ptitle{Correspondence Merging.}
After correcting the local correspondences, we merge them into global correspondences $\mathcal{{G}}^{t}$. 
It is worth noting that some local regions overlap with each other. Therefore, we utilize a hash table $\mathbb{H}$ to store the index of the original point cloud, facilitating the merging of local correspondences to prevent point duplication. With this strategy, we further enhance the efficiency of our method and save memory overhead. Ultimately, we obtain the global correspondences $\mathcal{{G}}^{t}=\bigcup_{i=1}^{n} \mathcal{{G}}_{i}^{t}$.

\ptitle{Global Correspondence Correction.}
Although we achieve the correspondence correction in local regions, if the seed correspondences $ \mathcal{\widetilde{G}}^{t-1}$ are significantly incorrect, it may still result in wrong refinement for the entire local region.
Therefore, similar to local correction, we 
refine global correspondences $\mathcal{{G}}^{t} \leftarrow \Omega_g (\mathcal{{G}}^{t})$, where $\Omega_g$ represents the operation of global correspondence correction. The key difference is that, in the global stage, we use second-order consistency~\cite{chen2022sc2} to identify correct correspondences.
By leveraging correspondences that satisfy maximum consistency, 
we rectify anomalous correspondences globally, completing one iteration of correspondence regeneration $\varTheta(\cdot)$.

\begin{table*}[htbp]
    \centering
     \scriptsize
      \resizebox{1.0\linewidth}{!}{
          \begin{tabular}{l|cccccc|cccccc|c}
              \toprule
              & \multicolumn{6}{c|}{\textbf{FPFH} (Traditional Descriptor)} & \multicolumn{6}{c|}{\textbf{FCGF} (Learning-based Descriptor)}  & \\
              Method & RR($\uparrow$) & RE($\downarrow$) & TE($\downarrow$) & IP($\uparrow$)  & INR($\uparrow$) & IN($\uparrow$) 
              & RR($\uparrow$) & RE($\downarrow$) & TE($\downarrow$) & IP($\uparrow$)  & INR($\uparrow$) & IN($\uparrow$) &   Time(s)      
              \\ \midrule 
              PointDSC~\cite{PointDSC}  &76.96	&2.25	&6.76	&66.64	&70.16	&290.23	
              &92.61	&2.10	&6.48	&78.37	&85.88	&1174.62   &0.22          \\ 
              VBReg~\cite{jiang2023robust}  &78.01 &2.28 &7.34 &67.23 &70.93 & 166.56  
              &92.77 &2.24 &6.82 &79.92 &86.30 &690.38 &1.02\\ 
              Hunter~\cite{yao2023hunter}  &\underline{84.70} &\underline{1.80} &\underline{6.46} &\underline{74.39} &\underline{79.27} & \underline{309.28} 
              &\textbf{94.05} &\underline{1.86} &6.54 &\underline{82.55} &\underline{88.01} &\underline{1203.88} &0.59\\ 
              \midrule
              SM~\cite{leordeanu2005spectral}  & 55.88 & 2.94 & 8.15 & 47.96 & 70.69 & 230.03
              & 86.57 & 2.29 & 7.07 & 81.44 & 38.36 &405.10 &\textbf{0.03}\\
              TEASER~\cite{yang2020teaser} & 75.48 & 2.48 & 7.31 &{73.01} &78.60  &270.98
              & 85.77 & 2.73 & 8.66 &{82.43} &87.90  &1182.44  & 0.07\\
              RANSAC-1M~\cite{fischler1981random} & 64.20 & 4.05 & 11.35  & 63.96  &69.03  &264.09 
              & 88.42 & 3.05 & 9.42 & 77.96 &83.08  &672.22 & 0.93 \\
              SC$^2$-PCR~\cite{chen2022sc2} &83.24	&2.21	&6.70	&73.64	&78.78	&	289.17
              &93.16	&2.11	&\underline{6.44}	&80.17	&87.24	&1140.85 &\underline{0.11}\\
              MAC~\cite{zhang20233d} &83.90 &2.11 &6.80 &- &-&- &93.72 &2.07 &6.52 &- &-&- &0.95\\
              FastMAC~\cite{zhang2024fastmac} &82.87 &2.15 &6.73 &- &-&- &92.67 &2.00 &6.47 &- &-&- &0.27\\
              Regor (\emph{ours})&\textbf{88.48}  &\textbf{1.70} &\textbf{5.94} &\textbf{82.68} &\textbf{1253.74} & \textbf{2532.40}   &\underline{93.96}  &\textbf{1.75} &\textbf{6.14} &\textbf{87.48} &\textbf{263.81} &\textbf{2620.30} &0.36
  
              \\
              \bottomrule
      \end{tabular}
      }
      \caption{Quantitative comparison on the {3DMatch} dataset with descriptors FPFH~\cite{FPFH} and FCGF~\cite{choy2019fully}.}
      \label{3dmatch2}
  \end{table*}

\section{Experiment}
We evaluate our algorithm using 
the indoor 3DMatch dataset~\cite{zeng20173dmatch} and the outdoor KITTI dataset~\cite{KITTIdataset}, 
demonstrating its superior performance. 
Furthermore, we introduce the 3DMatch-EOR benchmark and conduct experiments to demonstrate the effectiveness of our method in scenarios with very few correct initial correspondences.
Additionally, we 
also conduct robustness test experiments to verify that our 
algorithm can achieve robust registration using only weak features, 
thereby demonstrating its robustness across different descriptors. 
Finally, we conduct ablation studies to validate the effectiveness of each module.

\ptitle{Evaluation Metric.}
Following~\cite{chen2022sc2,zhang20233d}, 
we use registration recall (RR), rotation error (RE), and translation 
error (TE) to evaluate registration performance. For outdoor scenes, 
we consider a registration successful if the error is within 
(5$^{\circ}$, 60 cm), and for indoor scenes, within (15$^{\circ}$, 30 cm). 
Additionally, we assess the quality of correspondences using feature matching recall (FMR) and inlier precision (IP) 
which represents the proportion of inliers to the total correspondences. 
To further demonstrate our method's ability to generate a higher number of 
accurate correspondences, we define two new metrics: inlier number ratio (INR) 
and inlier numbers (IN). INR represents the ratio of the final number of 
inliers $\mathrm{IN}_{final}$ to the initial number of inliers $\mathrm{IN}_{initial}$,
defined explicitly in Appendix.

\subsection{Evaluation on Indoor Scenes}\label{Indoor1}

\ptitle{Experimental Setup.}
For indoor scenes, we utilize the 3DMatch dataset. 
Our method is validated on both 3DMatch~\cite{zeng20173dmatch} and 3DLoMatch benchmarks~\cite{huang2021predator}. 
Following~\cite{PointDSC,chen2022sc2}, we perform voxel 
downsampling on the raw point clouds, and use two descriptors, 
FPFH~\cite{FPFH} and FCGF~\cite{choy2020deep}, to extract features as inputs for each method. 

\begin{figure}[t]
    \centering{\includegraphics[width=1.0\linewidth]{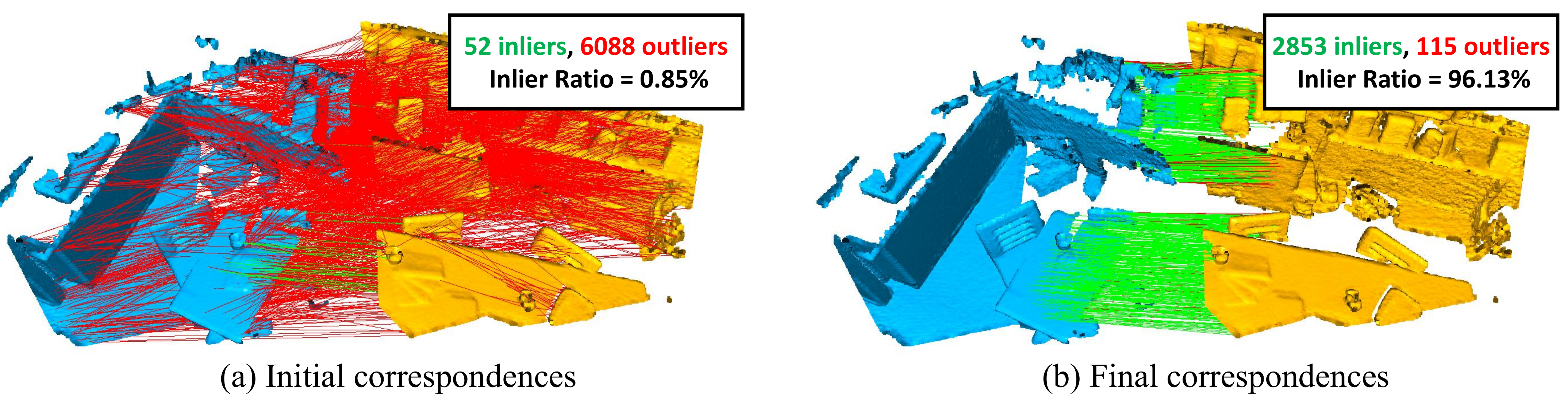}}
    \caption{Correspondences before and after processing.}
    \label{fig:processing}
\end{figure}

\begin{table}[t]
    \renewcommand{\arraystretch}{0.9}
    \centering
    \resizebox{1.0\linewidth}{!}{
    \begin{tabular}{l|cccccc}
    \toprule
    & {RR(\%)} & {RE($^{\circ}$)} & {TE(cm)} &{IP(\%)}  &{INR(\%)}  &{IN(/)} \\
    \midrule
    \multicolumn{7}{c}{\textbf{FPFH}}\\
    \midrule

    PointDSC~\cite{PointDSC} &27.91	&4.27	&10.45	&23.80	&25.76	&46.52\\
    VBReg~\cite{jiang2023robust} &30.83 &4.38 &10.92 &26.61 &29.70  & 48.41\\ 
    Hunter~\cite{yao2023hunter}  &36.90  &3.89   &10.05  &\underline{31.55}  &\underline{36.17}  &\underline{56.01}\\
    RANSAC~\cite{fischler1981random} &19.83	&4.67	&10.32	&30.82 &21.20	&37.02 \\
    SC$^2$-PCR~\cite{chen2022sc2}   &35.93	&4.26	&10.86	&30.11	&35.65	&55.23\\
    MAC~\cite{zhang20233d} &\underline{40.88} &\underline{3.66} &\underline{9.45}  &- &-&-\\
    FastMAC~\cite{zhang2024fastmac} &38.46 &4.04 &10.47 &- &-&-\\
    PCRegen (\emph{ours})  &\textbf{43.96}  &\textbf{2.89} &\textbf{8.93} &\textbf{37.73} &\textbf{778.35} &\textbf{643.24} 
    \\
    \midrule


    \multicolumn{7}{c}{\textbf{Predator}}\\
    \midrule
    PointDSC~\cite{PointDSC}  &68.30  &\underline{3.45}	&\textbf{9.44}	&56.35	&66.70	&863.40\\
    VBReg~\cite{jiang2023robust}     &69.82  &3.87  &10.07  &56.90  &67.09  &872.05\\ 
    Hunter~\cite{yao2023hunter}    &\underline{71.10}  &3.58  &9.72   &58.90  & \underline{69.06} &\underline{892.07}\\
    RANSAC~\cite{fischler1981random}  &64.85 &4.28 &11.04 &56.44 &66.81 & 869.21\\
    SC$^2$-PCR~\cite{chen2022sc2}    &69.46	&3.48	&9.63	&58.07	&68.63	&883.59\\
    MAC~\cite{zhang20233d} & 70.91 &3.69 &9.81 &- &-&-\\
    FastMAC~\cite{zhang2024fastmac} &68.77 &3.90 &10.32 &- &-&-\\
    PCRegen (\emph{ours})  &\textbf{72.04}  &\textbf{3.17} &\underline{9.46} &\textbf{63.95} &\textbf{220.46} &\textbf{2138.47}  \\
    \bottomrule
    \end{tabular}}
    \caption{
        Quantitative comparison on the 3DLoMatch Dataset.
    }
    \label{table:3DLoMatch}
\end{table}

\begin{figure*}[t]
    \centering{\includegraphics[width=1.0\textwidth]{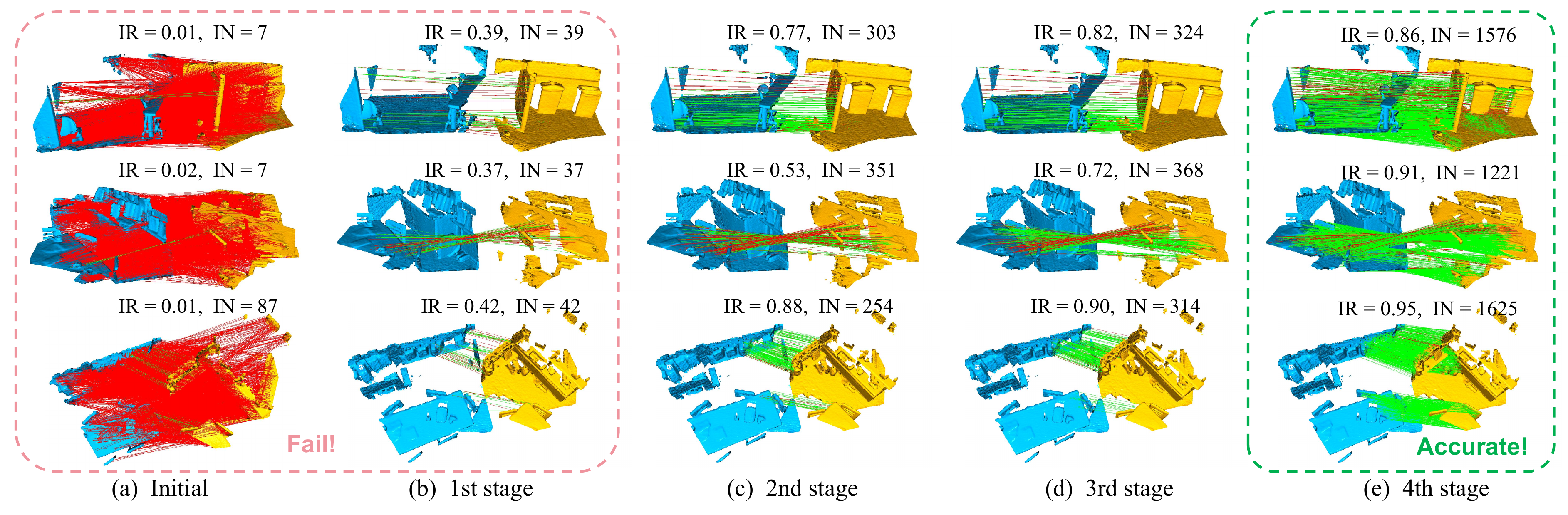}}  
    \caption{Correspondences visualization in different stages.}
    \label{fig:corr}
\end{figure*}

\begin{table*}[htbp]
    \scriptsize
    \renewcommand{\arraystretch}{0.9}
    \centering
    \resizebox{1.0\linewidth}{!}{
    \begin{tabular}{l|ccccccc|ccccccc}
    \toprule
    & {RR(\%)} & {RE($^{\circ}$)} & {TE(cm)} &{FMR(\%)} &{IP(\%)}  &{INR(\%)}  &{IN(/)}   & {RR(\%)} & {RE($^{\circ}$)} & {TE(cm)} &{FMR(\%)} &{IP(\%)}  &{INR(\%)}  &{IN(/)}\\

    \midrule
    &\multicolumn{14}{c}{\emph{Outlier ratio above 90\%}}\\
    \midrule
    & \multicolumn{7}{c|}{{3DMatch}} & \multicolumn{7}{c}{{3DLoMatch}}\\
    \midrule
    PointDSC~\cite{PointDSC} &71.12 &2.52 &7.79 &77.76 &60.05 &63.55 &129.12  
    &23.27 &4.84 &12.94 &39.78 &18.16 &26.57 &31.18\\
    SC$^2$-PCR~\cite{chen2022sc2}   &78.58	&2.49	&7.73	&84.02	&51.91		&{73.73}	&{138.98}  &35.39	&4.29	&10.96	&42.58	&29.6		&{35.14}	&{49.73}\\
    PCRegen (\emph{ours}) &\textbf{84.88} &\textbf{1.86} &\textbf{6.67} &\textbf{86.38} &\textbf{78.14} &\textbf{1457.54} &\textbf{2334.36}   &\textbf{43.50} &\textbf{2.95} &\textbf{9.24} &\textbf{46.34} &\textbf{36.96} &\textbf{780.05} &\textbf{632.62}\\
    
    \midrule
    &\multicolumn{14}{c}{\emph{Outlier ratio above 99\%}}\\
    \midrule
    & \multicolumn{7}{c|}{{3DMatch}} & \multicolumn{7}{c}{{3DLoMatch}}\\
    \midrule
    PointDSC~\cite{PointDSC} &2.56 &5.12 &15.28 &10.90 &3.07 &4.88 &0.17      &0.75 &4.93 &29.34 &0.75 &0.08 &1.46 &0.02\\
    SC$^2$-PCR~\cite{chen2022sc2}   &9.15	&4.94	&12.21	&16.20	&6.08  &{9.61}	&{0.34} &3.28	  &5.94	&19.44	&7.88	&2.47	&{4.96}	&{0.89}\\
    PCRegen (\emph{ours}) &\textbf{26.76} &\textbf{2.17} &\textbf{8.60} & \textbf{28.87} &\textbf{21.54} &\textbf{1545.12} &\textbf{666.94}    &\textbf{9.27} &\textbf{3.36} &\textbf{12.39} &\textbf{11.84} &\textbf{6.30} &\textbf{291.40} &\textbf{118.45}\\
    
    \bottomrule

    \end{tabular}
    }
    \caption{
        Results on the 3DMatch-EOR benchmark with extreme outlier ratios.
    }
    \label{table:extreme}
\end{table*}

\ptitle{Result on 3DMatch.}
We compare our method against several baselines on the 3DMatch dataset, 
including PointDSC~\cite{PointDSC}, 
VBReg~\cite{jiang2023robust}, Hunter~\cite{yao2023hunter}, SM~\cite{leordeanu2005spectral}
TEASER~\cite{yang2020teaser}, RANSAC~\cite{fischler1981random}, 
SC$^2$-PCR ~\cite{chen2022sc2}, MAC~\cite{zhang20233d}, and 
FastMAC~\cite{zhang2024fastmac}. The first three baselines are learning-based methods, 
while others are geometry-based approaches. 
The results are summarized in Table~\ref{3dmatch2}.
Our method outperforms the previous state-of-the-art~\cite{chen2022sc2, zhang20233d}. 
Using the FPFH~\cite{choy2019fully}, our method outperforms all others across all metrics. Notably, our method 
achieves 88.48\% RR, 
surpassing SC$^2$-PCR~\cite{chen2022sc2} by 5.2 percentage points (pp).
It also achieves more accurate registration with the lowest RE and TE. 
One significant advantage 
of our method is obtaining more high-quality correspondences, as reflected in the INR and IN. 
Our Regor achieves a 1253.74\% INR, 
which is much better than 78.78\% of SC2-PCR. 
Previous methods~\cite{chen2022sc2, zhang2024fastmac, zhang20233d} only perform top-down filtering, which inevitably limits the INR to below 100\%. In contrast, our Regor regenerates more high-quality correspondences, producing around 10 times the original amount. This advantage allows our method to achieve robust registration even when there are few initial correct correspondences. Moreover, despite employing multiple iterations, it still maintains a comparable speed to the baselines~\cite{jiang2023robust, yao2023hunter} due to its simple sampling strategy and efficient consistency computation.

\ptitle{Result on 3DLoMatch.}
Since Predator~\cite{huang2021predator} is specifically designed for scenarios with low overlap, we use Predator~\cite{huang2021predator} and FPFH~\cite{FPFH} to extract features on 3DLoMatch.
The comparative results are presented in Table~\ref{table:3DLoMatch}. 
Our approach achieves the highest RR and RE on 3DLoMatch, regardless of the descriptors used.
Due to the inherent weakness of FPFH, nearly all baselines perform poorly on 3DLoMatch. 
In contrast, our method shows significant performance improvement, 
surpassing MAC~\cite{zhang20233d} by 3.08 pp. 

\ptitle{Result on 3DMatch-EOR.}
The key to the effectiveness of our method lies in its ability to generate more accurate correspondences, particularly in scenarios with extremely high outlier ratios. To validate this, we select extreme outlier ratio cases from the 3DMatch/3DLoMatch datasets and propose the 3DMatch-EOR benchmark which is described in Appendix. We establish two benchmarks with outlier rates exceeding 99\% and 90\%, respectively. Our method is compared to SC2-PCR~\cite{chen2022sc2} and PointDSC~\cite{PointDSC}, with results shown in Table~\ref{table:extreme}. In cases with such high outlier rates, achieving robust registration is exceptionally challenging, as the number of initial correct correspondences is minimal. When the outlier rate exceeded 99\%, the average inlier count of SC2-PCR is less than 1, leading to registration failure and achieving only 9.15\%/3.28\% in RR. In contrast, our method achieves an average of 667 inliers, an improvement of nearly 2000 times compared to SC2-PCR~\cite{chen2022sc2}, resulting in a 26.76\% RR. Our Regor enables success in scenarios where failure would otherwise be inevitable,
allowing for relatively robust registration even with few initial inliers.

\subsection{Evaluation on Outdoor Scenes}\label{Outdoor}

\ptitle{Experimental Setup.}
Following~\cite{chen2023sc2,zhang20233d}, 
we validate our method's performance on the outdoor KITTI dataset. 
We select sequences  8 to 10 for testing. Consistent with the settings~\cite{chen2022sc2}, 
we perform downsampling using a 30 cm voxel grid and extract features using FPFH descriptors.

\begin{table}[htbp]
    \renewcommand{\arraystretch}{0.9}
    \centering
    \resizebox{1.0\linewidth}{!}{
    \begin{tabular}{l|cccccc}
    \toprule
    & {RR(\%)} & {RE($^{\circ}$)} & {TE(cm)} &{IP(\%)}  &{INR(\%)}  &{IN(/)} \\
    \midrule
    PointDSC~\cite{PointDSC}   &99.46	&0.57	&7.22	&91.39	&93.05	&247.11\\
    VBReg~\cite{jiang2023robust}   &98.92 &8.39  &8.41  &90.82 &92.13 &246.09\\
    Hunter~\cite{yao2023hunter}  &\textbf{99.82} &\underline{0.55}  &\underline{6.29}  &\underline{92.11} &94.58 & 247.50\\
    RANSAC~\cite{fischler1981random}   &74.41 &1.55  &30.20  &85.29 &87.20 &189.70 \\
    SC$^2$-PCR~\cite{chen2022sc2}   &\textbf{99.82}	&0.61	&8.16	&\textbf{93.59} &\underline{95.94}	&\underline{248.46}\\
    MAC~\cite{zhang20233d}   &99.46	&0.59	&8.70 &- &- &-   \\
    FastMAC~\cite{zhang2024fastmac} &98.02 &\underline{0.55} &8.24  &- &- &-\\
    Regor (\emph{ours}) &\textbf{99.82}	&\textbf{0.50}	&\textbf{5.36}	&79.57	&\textbf{2486.82}	&\textbf{5353.22}\\
    \bottomrule

    \end{tabular}
    }
    \caption{
        Quantitative comparison on the KITTI Dataset. 
    }
    \label{table:kitti}
\end{table}

\ptitle{Result on KITTI.}
We conduct comparative experiments between our method and
PointDSC~\cite{PointDSC}, VBReg~\cite{jiang2023robust}, Hunter~\cite{yao2023hunter}, 
RANSAC~\cite{fischler1981random}, SC$^2$-PCR~\cite{chen2022sc2}, 
MAC~\cite{zhang20233d}, and FastMAC~\cite{zhang2024fastmac} on the KITTI dataset. 
The results are presented in Table~\ref{table:kitti}. 
Our Regor achieves the best performance with the highest RR, outperforming RANSAC~\cite{fischler1981random} by 25.41 pp. 
It also demonstrates 
lower RE and TE, achieving the most accurate registration. 
Furthermore, our method shows superior performance in generating new correspondences, 
with INR over 25 times higher than those of other baselines~\cite{yao2023hunter,chen2022sc2}.
Despite IP is lower compared to SOTA~\cite{chen2022sc2}, 
our method achieves more accurate pose estimation
due to a larger number of high-quality inliers.

\begin{figure}[t]
    \centering{\includegraphics[width=1.0\linewidth]{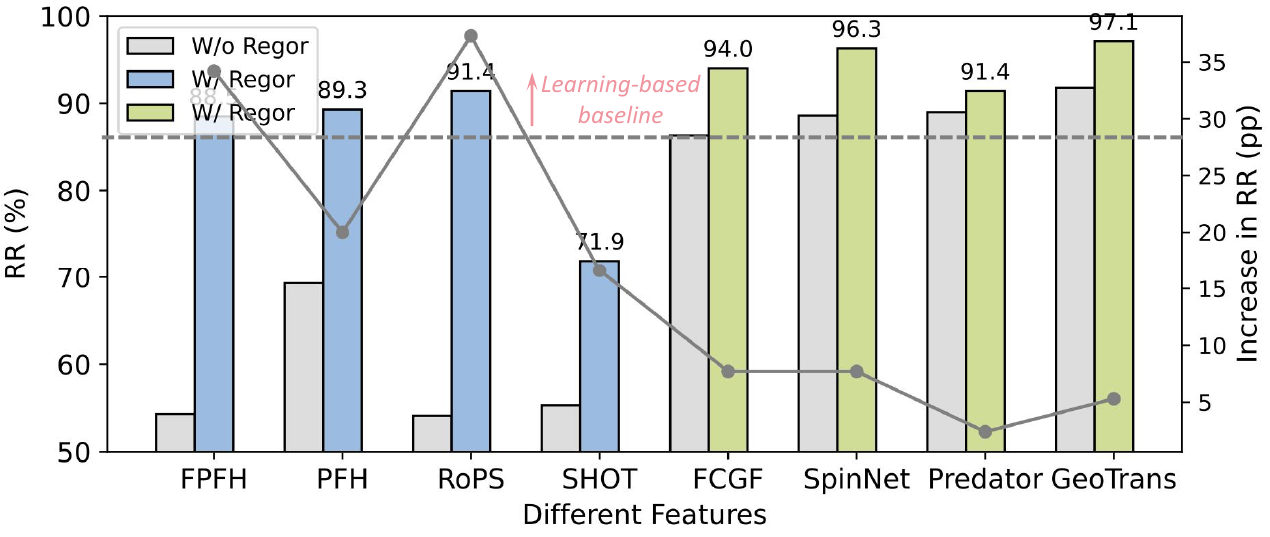}}
    \caption{RR under different features.}
    \label{fig:Features}
\end{figure}

\begin{table}[t]
    \setlength{\tabcolsep}{1.5pt}
    \centering
    \resizebox{1.0\linewidth}{!}{
    \begin{tabular}{l|lll}
    \toprule
    Method & RR(\%)$\uparrow$  & IP(\%)$\uparrow$ & IN(/)$\uparrow$ \\
    \midrule
    FPFH~\cite{FPFH} &54.34  & 45.74  &228.71   \\
    FPFH~\cite{FPFH} + Regor &88.48 (+\textbf{34.14}) &82.68 (+\textbf{36.94}) &1253.74 ($\times$\textbf{5.48})  \\
    \hline
    PFH~\cite{PFH}   &69.38 &57.28 &286.38   \\
    PFH~\cite{PFH} + Regor &89.28 (+\textbf{19.90}) &81.72 (+\textbf{24.44}) &2477.39 ($\times$\textbf{8.65}) \\
    \hline
    RoPS~\cite{guo2013rotational}   &54.16 &42.49 &212.46  \\
    RoPS~\cite{guo2013rotational} + Regor &91.37 (+\textbf{37.21}) &84.29 (+\textbf{41.80}) &2542.32 ($\times$\textbf{11.96})  \\
    \hline
    SHOT~\cite{salti2014shot}   & 55.33&46.01 &229.50   \\
    SHOT~\cite{salti2014shot}  + Regor &71.90 (+\textbf{16.57}) &66.41 (+\textbf{20.40}) &1981.10 ($\times$\textbf{8.63}) \\
    \hline
    FCGF~\cite{choy2019fully} &86.32   & 56.85  &405.27  \\
    FCGF~\cite{choy2019fully} + Regor   &93.96 (+\textbf{7.64}) &87.48 (+\textbf{30.63}) &2620.30 ($\times$\textbf{6.46})   \\
    \hline
    SpinNet~\cite{ao2021spinnet} &88.60    &47.53  &444.84  \\
    SpinNet~\cite{ao2021spinnet} + Regor &96.30 (+\textbf{7.70}) &88.24 (+\textbf{40.71}) &2799.74 ($\times$\textbf{6.29})   \\
    \hline
    Predator~\cite{huang2021predator} &89.00    &58.01  &444.38 \\
    Predator~\cite{huang2021predator} + Regor  &91.44 (+\textbf{2.44}) &86.84 (+\textbf{28.83}) &4608.34 ($\times$\textbf{10.37})   \\
    \hline
    GeoTrans~\cite{qin2022geometric}  &91.80 &75.20 &449.85   \\
    GeoTrans~\cite{qin2022geometric} + Regor  &\textcolor{red}{97.10} (+\textbf{5.30}) &91.01 (+\textbf{15.81}) &2649.30 ($\times$\textbf{5.88})   \\
    \bottomrule
    \end{tabular}
    }
    \caption{
        Enhancement of feature-based methods.
    }
    \label{Henhancement}
 \end{table}

\subsection{Robustness to Different Features}
To further assess the robustness of various descriptors, 
we also test with traditional descriptors such as 
FPFH~\cite{FPFH}, SHOT~\cite{salti2014shot}, PFH~\cite{PFH}, and PoRS~\cite{guo2013rotational}, 
as well as learning-based descriptors like SpinNet~\cite{ao2021spinnet}, 
FCGF~\cite{choy2019fully},
Predator~\cite{huang2021predator}, 
and GeoTrans~\cite{qin2022geometric}. As shown in Table~\ref{Henhancement}, 
our Regor consistently improves performance across all features, which demonstrates it is robust to different features.
Notably, GeoTrans~\cite{qin2022geometric}+Regor achieves a SOTA RR of 97.10\%.

\ptitle{Good Performance with Weak Features.}
Moreover, our method significantly enhance 
the performance of the weak descriptors,
increasing the RR from 54.16\% to 91.37\% with RoPS~\cite{guo2013rotational}.
As can be seen in Figure~\ref{fig:Features}, with our Regor, the traditional descriptors~\cite{PFH, guo2013rotational, FPFH} achieves 88\%+ RR, surpassing the learning-based baseline~\cite{choy2019fully}.
When combined with our Regor, 
RoPS~\cite{guo2013rotational} exhibits performance that exceeds the learning-based methods~\cite{choy2019fully,ao2021spinnet,huang2021predator}, only second to GeoTrans~\cite{qin2022geometric}.
This demonstrates that our method can achieve robust registration even with weak features. 



\subsection{Ablation Study}\label{sec.Ablation}


\begin{table}[htbp]
    \centering
    \resizebox{1.0\linewidth}{!}{
    \begin{tabular}{l|l|ccc|ccc}
    \toprule
    & &\multicolumn{3}{c|}{3DMatch} & \multicolumn{3}{c}{3DMatch-EOR} \\
    No.  &Methods        & RR  & IP& IN& RR & IP & IN \\
    \midrule
    1)         &One-stage   &86.81 &76.34 &290.32  &23.94 &15.88 &9.13 \\
    2)         &Progressive*  &\textbf{88.48} &\textbf{82.68}  &\textbf{2532.40}  &\textbf{26.76}  &\textbf{21.54}  &\textbf{666.94} \\
    \midrule
    3)         &NNM  &87.35&80.90&2487.23 &25.06 &19.82 &643.45\\
    4)         &MM  &86.81 &81.03 &2392.53 &23.94  &20.80 &600.09               \\
    5)         &GMM*  &\textbf{88.48} &\textbf{82.68}  &\textbf{2532.40}  &\textbf{26.76}  &\textbf{21.54}  &\textbf{666.94}  \\
    \midrule
    6)         &Only Local    &87.35  &78.40  &2526.63  &23.06  &17.32 &\textbf{668.08} \\ 
    7)         &Only Global  &76.37&80.48 &2502.21  &24.65 &16.57 &641.42 \\
    8)         &Local \& Global*  &\textbf{88.48} &\textbf{82.68}  &\textbf{2532.40}  &\textbf{26.76}  &\textbf{21.54}  &{666.94} \\
    \midrule
    9)         &SC &86.88 &81.09  &2491.54   &22.54&18.18&574.74      \\
    10)         &CTC*   &\textbf{88.48} &\textbf{82.68}  &\textbf{2532.40}  &\textbf{26.76}  &\textbf{21.54}  &\textbf{666.94}\\
    \bottomrule
    \end{tabular}
    }
    \caption{Ablation study on 3DMatch and 3DMatch-EOR. }
    \label{Ablation}
\end{table}

We conduct ablation experiments 
on the 3DMatch and 3DMatch-EOR benchmarks using FPFH descriptor. 

\ptitle{Progressive or One-stage.}
We verify the effectiveness of our progressive strategy by only using the one-stage process. The results are
shown in Table~\ref{Ablation} (1) and (2). Our strategy improves RR by 1.65/2.82 pp compared 
to the one-stage approach. 
It is attributed 
to the heuristic setting: 
starting with a larger radius ensures global optimality and robustness, 
and gradually reducing the radius with each iteration 
enhances accuracy.
More importantly, IN drops by 1$\sim$2 orders of magnitude, suggesting that our progressive strategy is key to generating more correspondences.

\ptitle{Matching Strategy.}
We replace our GMM with mutual matching (MM) and nearest neighbor matching (NNM).
Results are shown in Table~\ref{Ablation} (3), (4) and (5).
The model with GMM achieves the highest RR, IP, and IN.
Compared with MM, using GMM results in a significant improvement in IN, which indicates 
our GMM is helpful to generate more inliers
and addresses the issue of few initial inliers.

\ptitle{Local-Global Refinement.}
To demonstrate the importance of our correspondence refinement, 
we compare the complete model with two ablation models. 
The results, shown in Rows 6, 7, and 8 of Table ~\ref{Ablation}, indicate a 
significant performance drop when either local or global refinement is removed, 
with RR falling to 87\% and 86\%. In addition, the IP of the ablation model 
decreases by over 4 pp, which underscores the substantial 
contribution of the local-global refinement to the quality of the correspondences.


\ptitle{Spatial Compatibility.}
We compared the SOTA second-order consistency (SC)~\cite{chen2022sc2, chen2023sc2} with our CTC, and the results are shown at the bottom of Table~\ref{Ablation}. Our method with CTC obtains higher IP and IN, which indicates that CTC is more suitable for local refinement and achieves higher-quality correspondences. In addition, our CTC has less computational complexity than SC. Through it, we can implement an efficient iterative process.

\section{Conclusion}

In this paper, we introduce a new idea for optimizing correspondences. Unlike previous methods that focus solely on outlier removal, our Regor effectively addresses the challenge of limited correct initial correspondences. By generating more high-quality correspondences, our method enables more robust and accurate registration, achieving state-of-the-art performance across all datasets. Notably, it demonstrates significant improvements even under extreme outlier conditions. When combined with our Regor, using traditional descriptors can also achieve robust registration.


{\small
\bibliographystyle{ieeenat_fullname}
\bibliography{cvpr24}

\begin{thebibliography}{48}
\providecommand{\natexlab}[1]{#1}
\providecommand{\url}[1]{\texttt{#1}}
\expandafter\ifx\csname urlstyle\endcsname\relax
  \providecommand{\doi}[1]{doi: #1}\else
  \providecommand{\doi}{doi: \begingroup \urlstyle{rm}\Url}\fi

\bibitem[Ao et~al.(2021)Ao, Hu, Yang, Markham, and Guo]{ao2021spinnet}
Sheng Ao, Qingyong Hu, Bo Yang, Andrew Markham, and Yulan Guo.
\newblock Spinnet: Learning a general surface descriptor for 3d point cloud registration.
\newblock In \emph{CVPR}, pages 11753--11762, 2021.

\bibitem[Ao et~al.(2023)Ao, Hu, Wang, Xu, and Guo]{ao2023buffer}
Sheng Ao, Qingyong Hu, Hanyun Wang, Kai Xu, and Yulan Guo.
\newblock Buffer: Balancing accuracy, efficiency, and generalizability in point cloud registration.
\newblock In \emph{CVPR}, pages 1255--1264, 2023.

\bibitem[Bai et~al.(2021)Bai, Luo, Zhou, Chen, Li, Hu, Fu, and Tai]{PointDSC}
Xuyang Bai, Zixin Luo, Lei Zhou, Hongkai Chen, Lei Li, Zeyu Hu, Hongbo Fu, and Chiew-Lan Tai.
\newblock Pointdsc: Robust point cloud registration using deep spatial consistency.
\newblock In \emph{CVPR}, pages 15859--15869, 2021.

\bibitem[Barath and Matas(2018)]{barath2018graph}
Daniel Barath and Ji{\v{r}}{\'\i} Matas.
\newblock Graph-cut ransac.
\newblock In \emph{CVPR}, pages 6733--6741, 2018.

\bibitem[Barath and Valasek(2022)]{barath2022space}
Daniel Barath and G{\'a}bor Valasek.
\newblock Space-partitioning ransac.
\newblock In \emph{ECCV}, pages 721--737, 2022.

\bibitem[Chen et~al.(2022)Chen, Sun, Yang, and Tao]{chen2022sc2}
Zhi Chen, Kun Sun, Fan Yang, and Wenbing Tao.
\newblock Sc2-pcr: A second order spatial compatibility for efficient and robust point cloud registration.
\newblock In \emph{CVPR}, pages 13221--13231, 2022.

\bibitem[Chen et~al.(2023)Chen, Sun, Yang, Guo, and Tao]{chen2023sc2}
Zhi Chen, Kun Sun, Fan Yang, Lin Guo, and Wenbing Tao.
\newblock Sc2-pcr++: Rethinking the generation and selection for efficient and robust point cloud registration.
\newblock \emph{IEEE TPAMI}, 2023.

\bibitem[Choy et~al.(2019)Choy, Park, and Koltun]{choy2019fully}
Christopher Choy, Jaesik Park, and Vladlen Koltun.
\newblock Fully convolutional geometric features.
\newblock In \emph{ICCV}, pages 8957--8965, 2019.

\bibitem[Choy et~al.(2020)Choy, Dong, and Koltun]{choy2020deep}
Christopher Choy, Wei Dong, and Vladlen Koltun.
\newblock Deep global registration.
\newblock In \emph{CVPR}, pages 2514--2523, 2020.

\bibitem[Chua and Jarvis(1997)]{PointSignatures}
Chin~Seng Chua and Ray Jarvis.
\newblock Point signatures: A new representation for 3d object recognition.
\newblock \emph{IJCV}, 25:\penalty0 63--85, 1997.

\bibitem[Chum and Matas(2008)]{chum2008optimal}
Ond{\v{r}}ej Chum and Ji{\v{r}}{\'\i} Matas.
\newblock Optimal randomized ransac.
\newblock \emph{IEEE TPAMI}, 30\penalty0 (8):\penalty0 1472--1482, 2008.

\bibitem[Eppstein et~al.(2010)Eppstein, L{\"o}ffler, and Strash]{eppstein2010listing}
David Eppstein, Maarten L{\"o}ffler, and Darren Strash.
\newblock Listing all maximal cliques in sparse graphs in near-optimal time.
\newblock In \emph{Algorithms and Computation: 21st International Symposium}, pages 403--414, 2010.

\bibitem[Fischler and Bolles(1981)]{fischler1981random}
Martin~A Fischler and Robert~C Bolles.
\newblock Random sample consensus: a paradigm for model fitting with applications to image analysis and automated cartography.
\newblock \emph{Commun. ACM}, 24\penalty0 (6):\penalty0 381--395, 1981.

\bibitem[Gath and Geva(1989)]{gath1989unsupervised}
Isak Gath and Amir~B. Geva.
\newblock Unsupervised optimal fuzzy clustering.
\newblock \emph{IEEE TPAMI}, 11\penalty0 (7):\penalty0 773--780, 1989.

\bibitem[Geiger et~al.(2012)Geiger, Lenz, and Urtasun]{KITTIdataset}
Andreas Geiger, Philip Lenz, and Raquel Urtasun.
\newblock Are we ready for autonomous driving? the kitti vision benchmark suite.
\newblock In \emph{CVPR}, pages 3354--3361, 2012.

\bibitem[Gojcic et~al.(2019)Gojcic, Zhou, Wegner, and Wieser]{gojcic2019perfect}
Zan Gojcic, Caifa Zhou, Jan~D. Wegner, and Andreas Wieser.
\newblock The perfect match: 3d point cloud matching with smoothed densities.
\newblock In \emph{CVPR}, pages 5540--5549, 2019.

\bibitem[Guo et~al.(2013)Guo, Sohel, Bennamoun, Lu, and Wan]{guo2013rotational}
Yulan Guo, Ferdous Sohel, Mohammed Bennamoun, Min Lu, and Jianwei Wan.
\newblock Rotational projection statistics for 3d local surface description and object recognition.
\newblock \emph{IJCV}, 105:\penalty0 63--86, 2013.

\bibitem[Hu et~al.(2020)Hu, Yang, Xie, Rosa, Guo, Wang, Trigoni, and Markham]{hu2020randla}
Qingyong Hu, Bo Yang, Linhai Xie, Stefano Rosa, Yulan Guo, Zhihua Wang, Niki Trigoni, and Andrew Markham.
\newblock Randla-net: Efficient semantic segmentation of large-scale point clouds.
\newblock In \emph{CVPR}, pages 11108--11117, 2020.

\bibitem[Huang et~al.(2021{\natexlab{a}})Huang, Gojcic, Usvyatsov, Wieser, and Schindler]{huang2021predator}
Shengyu Huang, Zan Gojcic, Mikhail Usvyatsov, Andreas Wieser, and Konrad Schindler.
\newblock Predator: Registration of 3d point clouds with low overlap.
\newblock In \emph{CVPR}, pages 4267--4276, 2021{\natexlab{a}}.

\bibitem[Huang et~al.(2021{\natexlab{b}})Huang, Mei, Zhang, and Abbas]{huang2021comprehensive}
Xiaoshui Huang, Guofeng Mei, Jian Zhang, and Rana Abbas.
\newblock A comprehensive survey on point cloud registration.
\newblock \emph{arXiv preprint arXiv:2103.02690}, 2021{\natexlab{b}}.

\bibitem[Jiang et~al.(2023)Jiang, Dang, Wei, Xie, Yang, and Salzmann]{jiang2023robust}
Haobo Jiang, Zheng Dang, Zhen Wei, Jin Xie, Jian Yang, and Mathieu Salzmann.
\newblock Robust outlier rejection for 3d registration with variational bayes.
\newblock In \emph{CVPR}, pages 1148--1157, 2023.

\bibitem[Jin et~al.(2024)Jin, Armeni, Pollefeys, and Barath]{jin2024multiway}
Shengze Jin, Iro Armeni, Marc Pollefeys, and Daniel Barath.
\newblock Multiway point cloud mosaicking with diffusion and global optimization.
\newblock In \emph{CVPR}, pages 20838--20849, 2024.

\bibitem[Johnson and Hebert(1999)]{spinimage}
Andrew~E Johnson and Martial Hebert.
\newblock Using spin images for efficient object recognition in cluttered 3d scenes.
\newblock \emph{TPAMI}, 21\penalty0 (5):\penalty0 433--449, 1999.

\bibitem[Leordeanu and Hebert(2005)]{leordeanu2005spectral}
Marius Leordeanu and Martial Hebert.
\newblock A spectral technique for correspondence problems using pairwise constraints.
\newblock In \emph{ICCV}, pages 1482--1489, 2005.

\bibitem[Li and Hartley(2007)]{3D-3D}
Hongdong Li and Richard Hartley.
\newblock The 3d-3d registration problem revisited.
\newblock In \emph{ICCV}, pages 1--8. IEEE, 2007.

\bibitem[Lu et~al.(2021)Lu, Chen, Liu, Zhang, Qu, Liu, and Gu]{lu2021hregnet}
Fan Lu, Guang Chen, Yinlong Liu, Lijun Zhang, Sanqing Qu, Shu Liu, and Rongqi Gu.
\newblock Hregnet: A hierarchical network for large-scale outdoor lidar point cloud registration.
\newblock In \emph{ICCV}, pages 16014--16023, 2021.

\bibitem[Olsson et~al.(2008)Olsson, Kahl, and Oskarsson]{olsson2008branch}
Carl Olsson, Fredrik Kahl, and Magnus Oskarsson.
\newblock Branch-and-bound methods for euclidean registration problems.
\newblock \emph{IEEE TPAMI}, 31\penalty0 (5):\penalty0 783--794, 2008.

\bibitem[Pais et~al.(2020)Pais, Ramalingam, Govindu, Nascimento, Chellappa, and Miraldo]{3DRegNet}
G.~Dias Pais, Srikumar Ramalingam, Venu~Madhav Govindu, Jacinto~C. Nascimento, Rama Chellappa, and Pedro Miraldo.
\newblock 3dregnet: A deep neural network for 3d point registration.
\newblock In \emph{CVPR}, pages 7191--7201, 2020.

\bibitem[Qin et~al.(2022)Qin, Yu, Wang, Guo, Peng, and Xu]{qin2022geometric}
Zheng Qin, Hao Yu, Changjian Wang, Yulan Guo, Yuxing Peng, and Kai Xu.
\newblock Geometric transformer for fast and robust point cloud registration.
\newblock In \emph{CVPR}, pages 11143--11152, 2022.

\bibitem[Rusu et~al.(2008)Rusu, Blodow, Marton, and Beetz]{PFH}
Radu~Bogdan Rusu, Nico Blodow, Zoltan~Csaba Marton, and Michael Beetz.
\newblock Aligning point cloud views using persistent feature histograms.
\newblock In \emph{IROS}, pages 3384--3391, 2008.

\bibitem[Rusu et~al.(2009)Rusu, Blodow, and Beetz]{FPFH}
Radu~Bogdan Rusu, Nico Blodow, and Michael Beetz.
\newblock Fast point feature histograms (fpfh) for 3d registration.
\newblock In \emph{ICRA}, pages 3212--3217, 2009.

\bibitem[Salti et~al.(2014)Salti, Tombari, and Di~Stefano]{salti2014shot}
Samuele Salti, Federico Tombari, and Luigi Di~Stefano.
\newblock Shot: Unique signatures of histograms for surface and texture description.
\newblock \emph{CVIU}, 125:\penalty0 251--264, 2014.

\bibitem[Schnabel et~al.(2007)Schnabel, Wahl, and Klein]{schnabel2007efficient}
Ruwen Schnabel, Roland Wahl, and Reinhard Klein.
\newblock Efficient ransac for point-cloud shape detection.
\newblock In \emph{Comput. Graph Forum}, pages 214--226, 2007.

\bibitem[Xiong et~al.(2024)Xiong, Zheng, Xu, Wen, Shen, and Wang]{xiong2023speal}
Kezheng Xiong, Maoji Zheng, Qingshan Xu, Chenglu Wen, Siqi Shen, and Cheng Wang.
\newblock Speal: Skeletal prior embedded attention learning for cross-source point cloud registration.
\newblock In \emph{AAAI}, pages 6279--6287, 2024.

\bibitem[Yang et~al.(2022)Yang, Guo, Chen, and Tao]{yang2022one}
Fan Yang, Lin Guo, Zhi Chen, and Wenbing Tao.
\newblock One-inlier is first: Towards efficient position encoding for point cloud registration.
\newblock \emph{NeurIPS}, 35:\penalty0 6982--6995, 2022.

\bibitem[Yang et~al.(2020)Yang, Shi, and Carlone]{yang2020teaser}
Heng Yang, Jingnan Shi, and Luca Carlone.
\newblock Teaser: Fast and certifiable point cloud registration.
\newblock \emph{IEEE TRO}, 37\penalty0 (2):\penalty0 314--333, 2020.

\bibitem[Yang et~al.(2023)Yang, Zhang, Fan, Ren, and Zhang]{yang2023mutual}
Jiaqi Yang, Xiyu Zhang, Shichao Fan, Chunlin Ren, and Yanning Zhang.
\newblock Mutual voting for ranking 3d correspondences.
\newblock \emph{IEEE TPAMI}, 2023.

\bibitem[Yao et~al.(2023)Yao, Du, Cui, Ye, Wen, Zhang, Tian, and Gao]{yao2023hunter}
Runzhao Yao, Shaoyi Du, Wenting Cui, Aixue Ye, Feng Wen, Hongbo Zhang, Zhiqiang Tian, and Yue Gao.
\newblock Hunter: Exploring high-order consistency for point cloud registration with severe outliers.
\newblock \emph{IEEE TPAMI}, 2023.

\bibitem[Yi et~al.(2018)Yi, Trulls, Ono, Lepetit, Salzmann, and Fua]{CN-Net}
Kwang~Moo Yi, Eduard Trulls, Yuki Ono, Vincent Lepetit, Mathieu Salzmann, and Pascal Fua.
\newblock Learning to find good correspondences.
\newblock In \emph{CVPR}, pages 2666--2674, 2018.

\bibitem[Yin et~al.(2024)Yin, Xu, Lu, Chen, Xiong, Shen, Stachniss, and Wang]{yin2024survey}
Huan Yin, Xuecheng Xu, Sha Lu, Xieyuanli Chen, Rong Xiong, Shaojie Shen, Cyrill Stachniss, and Yue Wang.
\newblock A survey on global lidar localization: Challenges, advances and open problems.
\newblock \emph{IJCV}, pages 1--33, 2024.

\bibitem[Yu et~al.(2021)Yu, Li, Saleh, Busam, and Ilic]{yu2021cofinet}
Hao Yu, Fu Li, Mahdi Saleh, Benjamin Busam, and Slobodan Ilic.
\newblock Cofinet: Reliable coarse-to-fine correspondences for robust pointcloud registration.
\newblock \emph{NeurIPS}, 34:\penalty0 23872--23884, 2021.

\bibitem[Yu et~al.(2023{\natexlab{a}})Yu, Qin, Hou, Saleh, Li, Busam, and Ilic]{yu2023rotation}
Hao Yu, Zheng Qin, Ji Hou, Mahdi Saleh, Dongsheng Li, Benjamin Busam, and Slobodan Ilic.
\newblock Rotation-invariant transformer for point cloud matching.
\newblock In \emph{CVPR}, pages 5384--5393, 2023{\natexlab{a}}.

\bibitem[Yu et~al.(2023{\natexlab{b}})Yu, Ren, Zhang, Zhou, Lin, and Dai]{yu2023peal}
Junle Yu, Luwei Ren, Yu Zhang, Wenhui Zhou, Lili Lin, and Guojun Dai.
\newblock Peal: Prior-embedded explicit attention learning for low-overlap point cloud registration.
\newblock In \emph{CVPR}, pages 17702--17711, 2023{\natexlab{b}}.

\bibitem[Yu et~al.(2024)Yu, Qin, Zheng, and Xu]{yu2024learning}
Zhiyuan Yu, Zheng Qin, Lintao Zheng, and Kai Xu.
\newblock Learning instance-aware correspondences for robust multi-instance point cloud registration in cluttered scenes.
\newblock \emph{arXiv preprint arXiv:2404.04557}, 2024.

\bibitem[Zeng et~al.(2017)Zeng, Song, Nie{\ss}ner, Fisher, Xiao, and Funkhouser]{zeng20173dmatch}
Andy Zeng, Shuran Song, Matthias Nie{\ss}ner, Matthew Fisher, Jianxiong Xiao, and Thomas Funkhouser.
\newblock 3dmatch: Learning local geometric descriptors from rgb-d reconstructions.
\newblock In \emph{CVPR}, pages 1802--1811, 2017.

\bibitem[Zhang et~al.(2023)Zhang, Yang, Zhang, and Zhang]{zhang20233d}
Xiyu Zhang, Jiaqi Yang, Shikun Zhang, and Yanning Zhang.
\newblock 3d registration with maximal cliques.
\newblock In \emph{CVPR}, pages 17745--17754, 2023.

\bibitem[Zhang et~al.(2024)Zhang, Zhao, Li, and Chen]{zhang2024fastmac}
Yifei Zhang, Hao Zhao, Hongyang Li, and Siheng Chen.
\newblock Fastmac: Stochastic spectral sampling of correspondence graph.
\newblock In \emph{CVPR}, pages 17857--17867, 2024.

\bibitem[Zhao et~al.(2024)Zhao, Guo, Wang, and Ma]{zhao2023spherenet}
Guiyu Zhao, Zhentao Guo, Xin Wang, and Hongbin Ma.
\newblock Spherenet: Learning a noise-robust and general descriptor for point cloud registration.
\newblock \emph{IEEE TGRS}, 62:\penalty0 1--16, 2024.

\end{thebibliography}
}

\clearpage
\setcounter{page}{1}
\maketitlesupplementary

\section{Analysis and Proof}

\subsection{Proof of Theorem 1}
\ptitle{Theorem 1.}
\emph{
        Assuming event $\Phi $, where the consistency of correct correspondences is greater than the maximum consistency of incorrect correspondences, 
        has a probability ${P}(\Phi)\approx  1 $ (which means our method can correctly identify inliers),
        if the score of the local correspondences satisfies $Score(\mathcal{G}^{t}_i)\geqslant  1$, then the proportion of correct correspondences $b \geqslant a$.
}

\emph{Proof.}
    For simplicity, we denote the consistency matrix $\mathbf{S}^t_{i}$ as $\mathbf{S}_{\mathrm{CTC}}$. Thus, we can demonstrate that:
        \begin{equation}
       \begin{aligned} 
        &P(Score(\mathcal{G}^{t}_i)\geqslant1) = P(\frac{|| \mathbf{S}_{\mathrm{CTC}}||_1}{a\cdot N}\geqslant 1)=1 \\
        =& P(\frac{  \mathop{\max}\limits_{1\leqslant j \leqslant N} \sum_{i=1}^{N} |{\mathbf{S}_{\mathrm{CTC}}}_{ij}|   }{a\cdot N}\geqslant 1) \\
        =& P(\frac{  \mathop{\max}\limits_{1\leqslant j \leqslant N} \sum_{i=1}^{N} |{\mathbf{S}_{\mathrm{CTC}}}_{ij}|   }{a\cdot N}\geqslant 1 | \Phi) P(\Phi)\\
        +& P(\frac{  \mathop{\max}\limits_{1\leqslant j \leqslant N} \sum_{i=1}^{N} |{\mathbf{S}_{\mathrm{CTC}}}_{ij}|   }{a\cdot N}\geqslant 1 | \overline\Phi) P(\overline\Phi) \\
        \overset{(1)}{\approx}& P(\frac{  \mathop{\max}\limits_{1\leqslant j \leqslant N} \sum_{i=1}^{N} |{\mathbf{S}_{\mathrm{CTC}}}_{ij}|   }{a\cdot N}\geqslant 1 | \Phi) P(\Phi)\\
        =& \left[1- P(\frac{  \mathop{\max}\limits_{1\leqslant j \leqslant N} \sum_{i=1}^{N} |{\mathbf{S}_{\mathrm{CTC}}}_{ij}|   }{a\cdot N}< 1 | \Phi) \right]P(\Phi)\\
        =& \left[1- P(\frac{ \sum_{i=1}^{N} |{\mathbf{S}_{\mathrm{CTC}}}_{il}|   }{a\cdot N}< 1 | \Phi)^N \right]P(\Phi)\\
        \overset{(2)}{\leqslant }& \left[1- P(\frac{ b \cdot N   }{a\cdot N}< 1 )^N \right]P(\Phi)\\
        =& \left[1- P(\frac{b   }{a}< 1)^N \right]P(\Phi)\\
       \end{aligned} 
    \label{proof1}
    \end{equation}
where step (1) is based on our assumption that ${P}(\Phi)\approx  1 $. 
Step (2) can be easily derived 
from the definition of consistency matrix. By definition, the number of 
correct correspondences is $b\cdot N$. According to the definition of our consistency matrix, 
When the geometric consistency of correct correspondences is greater than the maximum consistency of incorrect correspondences, that is, event $\Phi$ occurs, there is always:
\begin{equation}
    \sum_{i=1}^{n} |{\mathbf{S}_{\mathrm{CTC}}}_{il}| \leqslant b \cdot N.
    \label{proof2}
\end{equation}
According to Eq.~\ref{proof2} and~\ref{proof1}, we have already proved that $\left[1- P(\frac{b   }{a}< 1)^N \right]P(\Phi) \geqslant 1$.
Below is our final derivation:
\begin{equation}
\begin{aligned} 
    &\left[1- P(\frac{b   }{a}< 1)^N \right]P(\Phi) \geqslant 1 \\
    \Rightarrow & \left[1- P(\frac{b   }{a}< 1)^N \right] \geqslant 1 \\
    \Rightarrow & P(\frac{b   }{a}< 1)^N \leqslant 0 \\
    \Rightarrow & P(\frac{b   }{a}< 1) \leqslant 0 \\
    \Rightarrow & P(\frac{b   }{a}\geqslant  1) =  1
\end{aligned} 
\end{equation}
Thus, we have proved the proportion of correct correspondences 
$b$ exceeds the threshold $a$.
Therefore, given $a$, we can distinguish local regions where the inlier ratio is greater than $a$.

This theorem is crucial for our correspondence refinement, as it allows control over correspondences through hyperparameter $a$ tuning. Additionally, It provides a theoretical guarantee for the quality of correspondence at each step. It is also valuable for other outlier removal methods.

\subsection{Proof of Theorem 2}
\ptitle{Theorem 2.}
\emph{
Our generalized mutual matching (GMM) generates at least as many correct correspondences as those produced by mutual matching (MM), that is, $|\mathcal{{G}}_{GMM}|\geqslant |\mathcal{{G}}_{MM}|$ where 
$\mathcal{{G}}_{GMM}$ and $\mathcal{{G}}_{MM}$ represent the correspondences obtained from the GMM and MM methods, respectively.  
}

\emph{Proof.}
According to the definition of the matching matrices $\mathbf{M}^{\mathcal{P}\rightarrow\mathcal{Q} }_1$ and $\mathbf{M}^{\mathcal{Q}\rightarrow\mathcal{P} }_1$ provided in Sec.~\ref{Grouping}, the mutual matching (MM) process can be expressed as follows:
\begin{equation}
    \mathcal{{G}}_{MM} = \mathbb{G}   \left( \mathbf{M}^{\mathcal{P}\rightarrow\mathcal{Q} }_1 \odot \mathbf{M}^{\mathcal{Q}\rightarrow\mathcal{P} }_1  \right) 
\end{equation}
where the function $\mathbb{G}(\mathbf{M})$ extracts correspondences from the matching matrix $\mathbf{M}$, which is defined as follows:
\begin{equation}
     \mathbb{G} \left( \mathbf{M}\right) = \left\{(\mathbf{p}_i, \mathbf{q}_j) | \mathbf{M}(i,j)=1, \mathbf{p}_i\in \mathbf{P}, \mathbf{q}_j\in \mathbf{Q}  \right\}
\end{equation}
In contrast, our GMM process can be expressed as:
\begin{equation}
    \mathcal{{G}}_{GMM} = \mathbb{G}   \left( \left( \mathbf{M}^{\mathcal{P}\rightarrow\mathcal{Q} }_1 \odot \mathbf{M}^{\mathcal{Q}\rightarrow\mathcal{P} }_2  \right) \otimes \left( \mathbf{M}^{\mathcal{Q}\rightarrow\mathcal{P} }_1 \odot \mathbf{M}^{\mathcal{P}\rightarrow\mathcal{Q} }_2  \right) \right) 
\end{equation}
Based on the definition an properties of function $\mathbb{G}$, we can derive the following:
\begin{equation}
    \begin{aligned} 
     & |\mathcal{{G}}_{GMM}|\\
     =&\left|\mathbb{G}   \left( \left( \mathbf{M}^{\mathcal{P}\rightarrow\mathcal{Q} }_1 \circ \mathbf{M}^{\mathcal{Q}\rightarrow\mathcal{P} }_2  \right) \otimes \left( \mathbf{M}^{\mathcal{Q}\rightarrow\mathcal{P} }_1 \circ \mathbf{M}^{\mathcal{P}\rightarrow\mathcal{Q} }_2  \right) \right) \right|\\
     \geqslant & \left|\mathbb{G}   \left( \left( \mathbf{M}^{\mathcal{P}\rightarrow\mathcal{Q} }_1 \circ \mathbf{M}^{\mathcal{Q}\rightarrow\mathcal{P} }_2  \right) \right) \right|\\
     \geqslant & \left|\mathbb{G}   \left( \left( \mathbf{M}^{\mathcal{P}\rightarrow\mathcal{Q} }_1 \circ \mathbf{M}^{\mathcal{Q}\rightarrow\mathcal{P} }_1  \right) \right) \right|\\
     =&|\mathcal{{G}}_{MM}|\\
    \end{aligned} 
 \label{proof4}
\end{equation}
 Thus, we complete the proof that  $|\mathcal{{G}}_{GMM}|\geqslant |\mathcal{{G}}_{MM}|$.

\subsection{Why our Regor perform well with few inliers?}
\begin{figure}[ht]
    \centering{\includegraphics[width=1.0\linewidth]{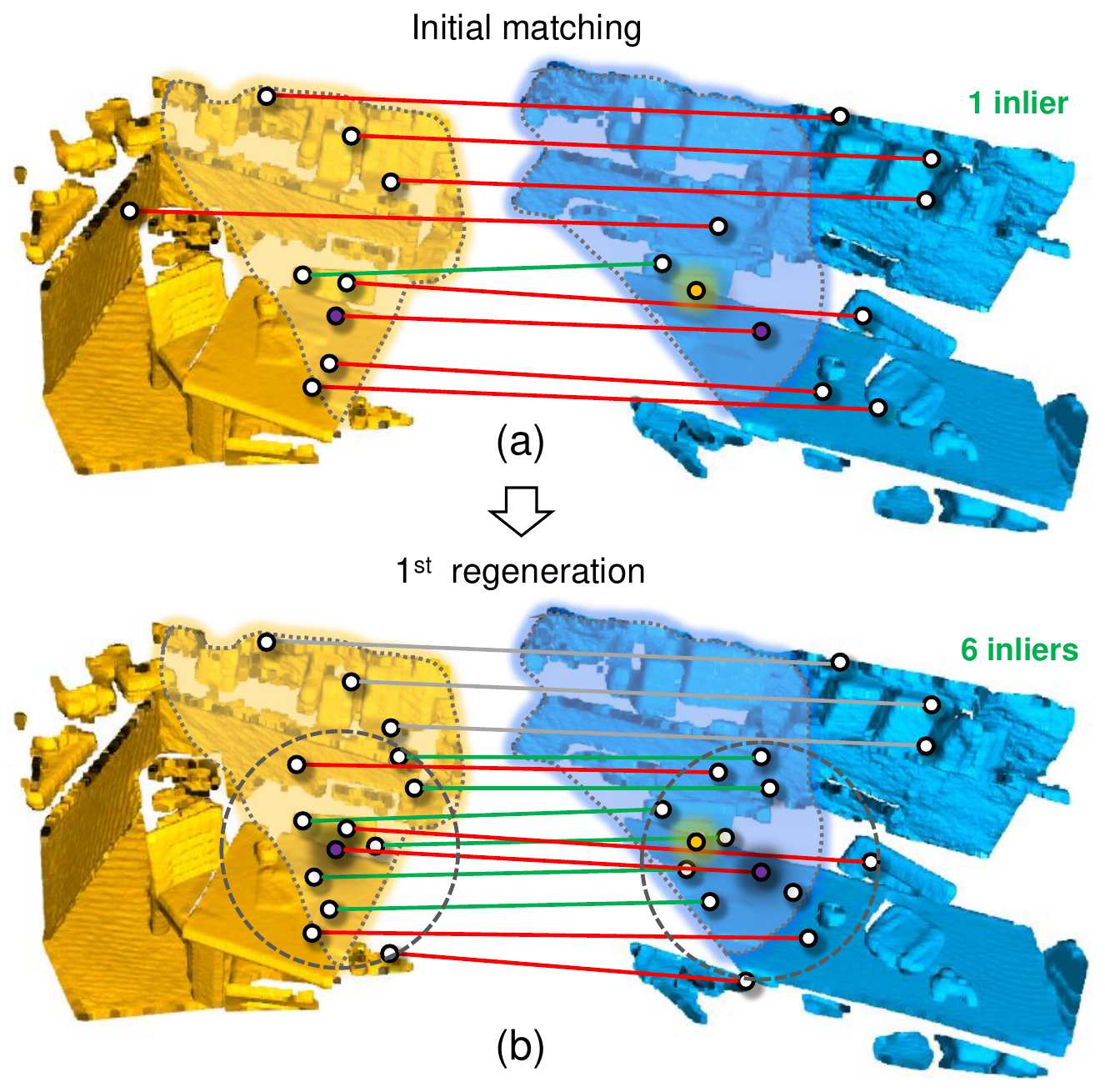}}  
    \caption{
        Illustration of our progressive correspondence regenerator.
        The green and red lines indicate correct and incorrect correspondences at the current stage, respectively. The purple and yellow dots represent the seed correspondences (center points) and really correct corresponding points, respectively.
        The luminous part is the overlapping area. 
        The dotted circles are the local corresponding regions.
       }
    \label{analysis}
\end{figure}

The advantage of our algorithm lies in its ability to achieve robust registration even with very few inliers or under extreme outlier rates, which is unattainable for existing outlier removal methods.
As shown in Fig.~\ref{analysis}a, the initial correspondences are filled with numerous incorrect matches, with only one inlier. Methods relying on geometric consistency, such as~\cite{chen2022sc2,zhang20233d}, would fail to extract this inlier.  Moreover, it is impossible to solve the 6DoF pose with only a single correct correspondence.

In contrast, our method still performs well in such harsh situations. The reasons can be summarized in the following two aspects. \textbf{First}, the whole process is gradually generated and corrected, which does not depend on the quality of the initial correspondence. \textbf{Second}, outliers with small errors conduce to regenerate new correct correspondences by local matching. 
For example, in Figure~\ref{analysis}, our approach begins by sampling a seed correspondence (the purple point in the figure).  For clarity, we illustrate a single seed point here.  Although the seed correspondence is incorrect (the correct match is the yellow point in the figure), our next step, local matching, is crucial.
If the seed correspondence has a relatively small distance error, it can still guide the matching of overlapping local regions.  In these regions, despite the initial error, the overlap enables the generation of new correct correspondences.  More importantly, performing local matching within these regions significantly reduces the search space compared to global matching across the entire point cloud, making it much easier to obtain correct correspondences.

Additionally, in the early iterations of our algorithm, we adopt a larger region radius for local matching.  This increases the probability of overlap between local regions, mitigating the impact of erroneous correspondences during the initial stages.  As the iterations progress, the region radius is gradually reduced to refine the matches and improve the overall quality of the correspondences.
As a result, even with very few inliers, our method can utilize erroneous correspondences with small errors to generate new correct correspondences, making accurate pose estimation possible under such challenging conditions.

\section{Implementation details}

\subsection{Point-level Pose Refinement}
Pose refinement is crucial for achieving accurate pose estimation. In the final stage of correspondence regeneration, we perform pose refinement based on the established correspondences. Unlike correspondence-level pose refinement such as SC2-PCR~\cite{chen2022sc2} and MAC~\cite{zhang20233d}, our approach conducts pose refinement at the point level.

A key advantage of our approach is its ability to generate dense and accurate correspondences. However, traditional pose refinement evaluation metrics, such as inlier count (IC)~\cite{chen2022sc2}, mean square error (MSE), truncated chamfer distance (TCD), and feature-and-spatial-consistency-constrained truncated chamfer distance (FS-TCD)~\cite{chen2023sc2}, rely heavily on initial correspondences. Given the limited number of correct initial correspondences, these metrics fail to fully capture the benefits of our dense and accurate correspondences. To address this limitation, we propose a novel point-level evaluation metric, termed point-level truncated chamfer distance (PO-TCD):
\begin{equation}
    { \mathbf{R}}^{*}, { \mathbf{t}}^{*}=\max _{\mathbf{R}, \mathbf{t}} 
    \sum_{\mathbf{{p}}_i \in \mathbf{{P}}} 
    \mathds {1}\left(
        \min _{\mathbf{{q}}_j \in \mathbf{{Q}}}
        \left\|  \mathbf{\mathbf{R} {p}}_i+\mathbf{t}-\mathbf{{q}}_j  \right\|< {\sigma_d}
        \right).
\end{equation}
It eliminates reliance on initial correspondences and instead computes the optimal truncation distance directly from dense point clouds, fully leveraging the benefits of our dense correspondences.

\subsection{Hyperparameter Settings}

\begin{table}[htbp]
    \centering
    \resizebox{1.0\linewidth}{!}{
    \begin{tabular}{l|cccccccc}
    \toprule
    Dataset  &$k_0$ &$r_0$ &$s_0$ &$\omega_k$ &$\omega_r$ &$\omega_s$ &$k_{GMM}$ &$a$\\
    \midrule
    3DMatch~\cite{zeng20173dmatch} &20 &1&500 &5 &0.5 &0.2 &3 &0.5   \\
    KITTI~\cite{KITTIdataset}   &20 &10 &500 &5 &0.5 &0.2 &3 &0.5   \\
    \bottomrule
    \end{tabular}
    }
    \caption{Hyperparameter settings in different datasets. }
    \label{Hyperparameters Settings}
\end{table}

The detailed hyperparameter settings are provided in Table~\ref{Hyperparameters Settings}. Due to the differing scales of indoor and outdoor datasets, we specify two distinct parameter configurations, differing only in the local region radius. Specifically, for indoor datasets, we set $r_0$ to 1m, while for outdoor datasets, $r_0$ is set to 10m. All other parameters remain consistent across both settings.

Here, we outline the implementation details of our approach. First, for local grouping, during the
$t$th iteration, we perform random sampling with sampling quantity
${s_0}\cdot{(w_s)^{t}}$ on the correspondences obtained from the $t-1$th iteration. Using a radius of ${r_0}\cdot{(r_s)^t}$, we construct local point sets and standardize the number of points in each local set to ${k_0}\cdot{(w_k)^{t}}$ by sampling to facilitate parallel computation. Next, we apply generalized mutual matching with 
$k_{GMM}=3$ to obtain reliable correspondences. Finally, through a refinement process that combines local and global correspondence optimization, the global correspondences are updated.
It is worth noting that before the first iteration of sampling, we utilize the SC2-PCR~\cite{chen2022sc2} module to ensure the initial quality of the correspondences. Additionally, after the final correspondence refinement stage, we perform point-level pose refinement to further enhance the results.

Throughout the entire process, point sets for each stage are stored and updated using index-based representations, significantly reducing memory overhead. Similarly, during Correspondence Merging, we employ an index-based hash $\mathbb{H}$ table to efficiently obtain unique global correspondences, ensuring both speed and memory efficiency.

\section{Detailed Metrics}

\ptitle{Inlier Number (IN). }
 It represents the number of correct correspondences in the final correspondences, reflecting our ability to generate inliers.

\ptitle{Inlier Number Ratio (INR).}
 It is the ratio of inliers $\mathrm{IN}_{a}$ in the final correspondences to the inliers $\mathrm{IN}_{b}$ in the initial correspondences, which can be calculated by:
\begin{equation}
    \mathrm{INR} = \frac{1}{H} \sum_{h=1}^H \mathrm{INR}_h,\,  \mathrm{INR}_h = 
    \begin{cases}
        \frac{\mathrm{IN}^{a}_h}{\mathrm{IN}^{b}_h}, \mathrm{IN}^{b}_h\neq 0\\
        \mathrm{IN}^{a}_h, \mathrm{IN}^{b}_h = 0
    \end{cases}
 \end{equation}
where $H$ represents the total number of point cloud pairs in the dataset, and 
$\mathrm{INR}_h$ denotes the INR for the $h$th point cloud pair.

\ptitle{Registration Recall (RR).}
Following~\cite{PointDSC,chen2022sc2}, we consider registration to be successful if the translation error (TE) and rotation error (RE) are within specified thresholds. For indoor scenes, the thresholds are $15^{\circ}$ and $30$cm, while for outdoor scenes, they are 
$5^{\circ}$ and $60$cm. The registration recall rate is calculated as:
\begin{equation}
    \mathrm{RR}=\frac{1}{H} \sum_{h=1}^H \mathds {1}\left(\mathrm{TE}_h<\sigma_d  \wedge  \mathrm{RE}_h<\sigma_\theta\right),
\end{equation}
where
$\mathds {1}$ represents the indicator function, and 
$\sigma_d$ and $\sigma_\theta $ are the thresholds for rotation error and translation error, respectively. $\mathrm{RE}_h$ and 
$\mathrm{TE}_h$ denote the rotation error and translation error for the 
for the $h$th point cloud pair. They can be computed using the following formulas:
\begin{equation}
    \begin{cases}
        \mathrm{RE}_h=\arccos \left(\frac{\operatorname{trace}\left(\hat{\mathbf{R}}_h^T \mathbf{R}_h\right)-1}{2}\right) \\
        \mathrm{TE}_h=\left\|\hat{\mathbf{t}}_h-\mathbf{t}_h\right\|
    \end{cases}
\end{equation}
where $\hat{\mathbf{R}}_h$, $\mathbf{R}_h$, $\hat{\mathbf{h}}_t$ and $\mathbf{h}_t$ are our predicted rotation matrix, the ground-truth rotation matrix,
predicted translation vector and the ground-truth translation vector, respectively.


\begin{table*}[htbp]
    \centering
    \resizebox{1.0\linewidth}{!}{
    \begin{tabular}{l|cccccc}
    \toprule
    Benchmark  &Indoor scene &Nuisances &Application scenarion &\# Matching pairs  \\
    \midrule
    3DMatch~\cite{zeng20173dmatch} &Indoor scene  &Occlusion, real noise &Registration &1623  \\
    3DLoMatch~\cite{huang2021predator} &Indoor scene  &Limited overlap, occlusion, real noise &Registration &1781  \\
    3DMatch-EOR 90\% (\emph{ours}) &Indoor scene  &High outlier ratio, occlusion, real noise &Registration &  1270\\
    3DMatch-EOR 99\% (\emph{ours}) &Indoor scene  &Extreme high outlier ratio, occlusion, real noise &Registration &  142\\
    KITTI~\cite{KITTIdataset} &Outdoor scene &Clutter, occlusion, real noise &Detection, registration, segmentation &555   \\
    \bottomrule
    \end{tabular}
    }
    \caption{Information of all tested datasets.}
    \label{tested datasets}
\end{table*}

\section{Additional Experiments}
The information of all tested datasets is given in Table~\ref{tested datasets}.

\subsection{Results at different numbers of inliers}

\begin{table*}[htbp]
    \centering
      \resizebox{1.0\linewidth}{!}{
          \begin{tabular}{cc|cccccc|cccccc}
              \toprule
              &  &\multicolumn{6}{c|}{Regor (\emph{ours})} & \multicolumn{6}{c}{Outlier removal~\cite{chen2022sc2}}  \\
              \# Inlier number &\# Pairs  &RR($\uparrow$) & RE($\downarrow$) & TE($\downarrow$) & IP($\uparrow$)  & INR($\uparrow$) & IN($\uparrow$) & RR($\uparrow$) & RE($\downarrow$) & TE($\downarrow$) & IP($\uparrow$)  & INR($\uparrow$) & IN($\uparrow$)     
              \\ \midrule 
                0$\sim$20   &55 & 3.87 &3.57  &10.36 &3.59 &273.01 &38.44
                &3.64 &4.82 &23.9 &1.49 &3.86 &0.02\\ 
                20$\sim$40  &95 &36.84  &3.98  &9.22 &29.65 &1510.95 &471.55
                &8.42 &5.16 &13.79  &10.33 &15.87 &0.29\\ 
                40$\sim$60   &80& 62.50  &2.71 &8.00 &52.83 &2536.91  &1275.64
                &27.5 &5.70 &12.86 &28.47 &28.47 &0.73\\ 
                60$\sim$80    &90 &80.00 &2.60 &8.50 &71.21 &2434.90 &1656.36   
                &63.33 &4.15 &11.62 &51.59 &61.64 &2.35\\ 
                80$\sim$100   &95 &91.58  &1.98 &8.04 &78.83 &2288.78 &2000.84 
                &86.32 &3.82  &11.22 &63.77 &72.58 &3.73\\ 
                100$\sim$200  &333 &95.50 &1.89 &7.00  &86.85 &1625.94 &2318.18 
                &92.19 &2.71  &8.24 &78.66 &85.15 &26.11\\ 
              \bottomrule
      \end{tabular}
      }
      \caption{Comparison between our Regor and outlier removal method~\cite{chen2022sc2} under different numbers of inliers. }
      \label{inlier_number}
\end{table*}

In Sec.~\ref{Indoor1}, we propose the 3DMatch-EOR Benchmark, which includes two baselines: one with an outlier ratio exceeding 99\% and another exceeding 90\%. From these experiments, we conclude that our method demonstrates greater robustness under extreme outlier ratios. However, since these experiments are designed based on ratios, they do not verify the impact of the number of inliers. To address this, we conduct additional comparative experiments using FPFH as the feature extraction baseline, varying the number of inliers. Specifically, we designed seven scenarios with inlier counts ranging from 0$\sim$20, 20$\sim$40, 40$\sim$60, 60$\sim$80, 80$\sim$100, to 100$\sim$200, comparing our method against the outlier removal method SC2-PCR~\cite{chen2022sc2}. The experimental results are summarized in Table~\ref{inlier_number}.

In all conditions, our method outperforms SC2-PCR~\cite{chen2022sc2} across all metrics. Even when the number of correct correspondences is extremely low (0$\sim$100), our method delivers excellent performance, thanks to its strong capability for generating new correspondences, as reflected in the INR and IN metrics. Furthermore, 
as the number of inliers decreases, feature matching becomes increasingly challenging, leading to a gradual decline in the RR of our method. However, the rate of decline for our method is significantly slower than that of SC2-PCR~\cite{chen2022sc2}, indicating a certain degree of robustness in handling scenarios with very few inliers. However, when IN is less than 20, our method also cannot achieve robust registration.

\subsection{Detailed Ablation Study}

\begin{figure}[ht]
    \centering{\includegraphics[width=1.0\linewidth]{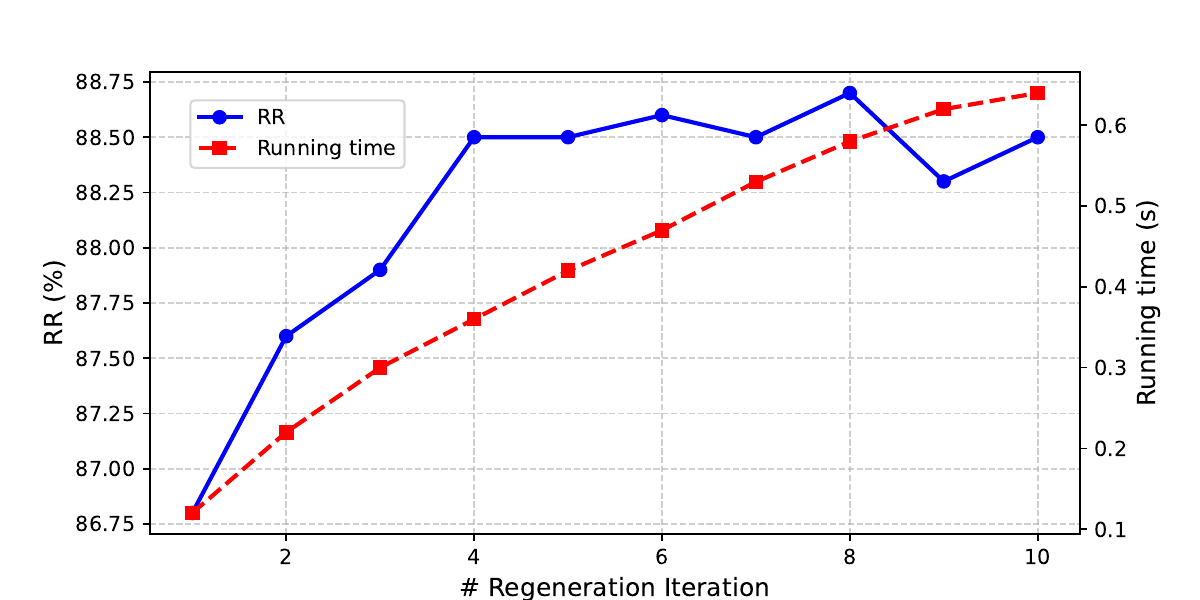}}  
    \caption{
        Ablation study on the progressive iteration.
       }
    \label{iteration}
\end{figure}

\ptitle{Progressive Iteration.}
In Section~\ref{sec.Ablation}, we conduct an ablation study to compare progressive and one-stage approaches. To further analyze the effect of iterations, we vary the number of iterations from 0 to 10 and evaluate the performance on the 3DMatch dataset. To balance registration success and speed, our study reports two metrics: registration recall (RR) and running time. The results, presented in Figure~\ref{iteration}, show that RR increases significantly as the number of iterations rises from 1 to 4. Beyond 4th iterations, the improvement slows, eventually saturating, with occasional fluctuations. Meanwhile, runtime increases steadily with more iterations. To balance performance and efficiency, we ultimately select 4 iterations as the optimal configuration.

\begin{table}[htbp]
    \centering
    \resizebox{1.0\linewidth}{!}{
    \begin{tabular}{l|ccc|ccc}
    \toprule
    &\multicolumn{3}{c|}{3DMatch} & \multicolumn{3}{c}{3DMatch-EOR} \\
    Methods        & RR  & IP& IN& RR & IP & IN \\
    \midrule
    1) K nearest neighbor  &88.32 &{82.55} &2522.09  &26.70 &21.40 &\textbf{670.23}  \\
    2) Radius nearest neighbor*  &\textbf{88.48} &82.68 &\textbf{2532.40} &\textbf{26.76} &\textbf{21.54} &{666.94}\\
    \midrule
    3) Spectral technique~\cite{leordeanu2005spectral,chen2022sc2}  &88.36 &\textbf{84.77} &2501.41  &26.19 &\textbf{24.33} &630.92\\
    4) Sampling with  consistency score   &88.06 &84.09 & 2510.91   &20.81 &22.19 &578.09 \\
    5) Random sampling*    &\textbf{88.48} &82.68 &\textbf{2532.40} &\textbf{26.76} &{21.54} &\textbf{666.94}\\
    \bottomrule
    \end{tabular}
    }
    \caption{Ablation study on prior-guided local grouping. }
    \label{grouping}
\end{table}

\ptitle{Prior-guided Local Grouping.}
We conduct a more detailed ablation study on the prior-guided local grouping module, focusing on two aspects: sampling strategies and neighbor grouping strategies. For neighbor grouping, we evaluate two approaches: K-nearest neighbors (KNN) and radius-based neighbors (RNN). For sampling, we test three methods: spectral technique~\cite{leordeanu2005spectral}, sampling with consistency scores, and random sampling.
As shown in Table~\ref{grouping}, the RNN strategy outperforms KNN. 
This is because RNN defines local region sizes independently of the point cloud's sparsity, while KNN is highly affected by variations in point cloud density. Surprisingly, as seen in Rows 3, 4, and 5 of Table~\ref{grouping}, random sampling outperforms the other strategies, achieving the highest RR, albeit with relatively lower IP. This is because random sampling promotes a more uniform distribution of local regions across the point cloud, reducing the risk of getting trapped in local optima and improving robustness. Meanwhile, ablated models focus local regions on areas with higher-quality correspondences as determined by our method, resulting in higher IP.


\begin{table}[htbp]
    \centering
    \resizebox{1.0\linewidth}{!}{
    \begin{tabular}{l|ccc|ccc}
    \toprule
    &\multicolumn{3}{c|}{3DMatch} & \multicolumn{3}{c}{3DMatch-EOR} \\
    Models        & RR  & IP& IN& RR & IP & IN \\
    \midrule
    1) $k_0=50$, $\omega_k=5$  &88.42 &80.48 &\textbf{2601.12}  &24.20 &20.11 &\textbf{681.23}  \\
    2) $k_0=20$, $\omega_k=10$  &\textbf{88.48} &81.82 &{2580.45} &{26.60} &{20.90} &{678.14}\\
    3) $k_0=20$, $\omega_k=5$*  &\textbf{88.48} &82.68 &{2532.40} &\textbf{26.76} &\textbf{21.54} &{666.94}\\
    \midrule
    4) $r_0=2$, $\omega_r=0.5$ &88.42 &84.09 & \textbf{2598.01}   &26.44 &20.91 &\textbf{702.29} \\
    5) $r_0=0.5$, $\omega_r=0.5$    &87.21&80.92 & 2423.12   &22.13 &21.49 &590.77 \\
    6) $r_0=1$, $\omega_r=0.2$   &87.43 &82.56 & 2499.86   &22.81 &21.88 &553.51 \\
    7) $r_0=1$, $\omega_r=0.5$*    &\textbf{88.48} &82.68 &{2532.40} &\textbf{26.76} &\textbf{21.54} &{666.94}\\
    \bottomrule
    \end{tabular}
    }
    \caption{Ablation study with different hyperparameters. }
    \label{Ablation:hyperparameters}
\end{table}

\ptitle{Hyperparameter Setting.}
We conduct an ablation study on hyperparameter settings using the 3DMatch dataset. To simplify hyperparameter tuning, we design the required sampling number $s_t$, the local region radius $r_t$, and the number of points in the local region $k_t$ to follow a geometric progression. This approach allows all parameters to be determined by setting only the initial values $s_0$, $r_0$, $k_0$, and the common ratios $w_s$, $w_r$, $w_k$. The results of the ablation study are presented in Table~\ref{Ablation:hyperparameters}. As shown in Rows 2, 3, 4, and 7 of Table~\ref{Ablation:hyperparameters}, increasing $k_0$ and $r_0$ leads to a higher number of inliers (IN), but does not necessarily result in a higher RR. Balancing accuracy and efficiency, we determine $k_0=20$ and $\omega_k=5$ based on Rows 1, 2, and 3. Similarly, Rows 4, 5, 6, and 7 indicated that $r_0=1$ and $\omega_r=0.5$ are the optimal radius hyperparameters.

\subsection{Additional Qualitative Results}
As demonstrated in Sec.~\ref{Indoor1} and \ref{Outdoor}, our Regor achieves superior quantitative performance on the 3DMatch and KITTI datasets. In this section, we provide additional visualizations to illustrate our qualitative results.

\begin{figure*}[ht]
    \centering{\includegraphics[width=1.0\linewidth]{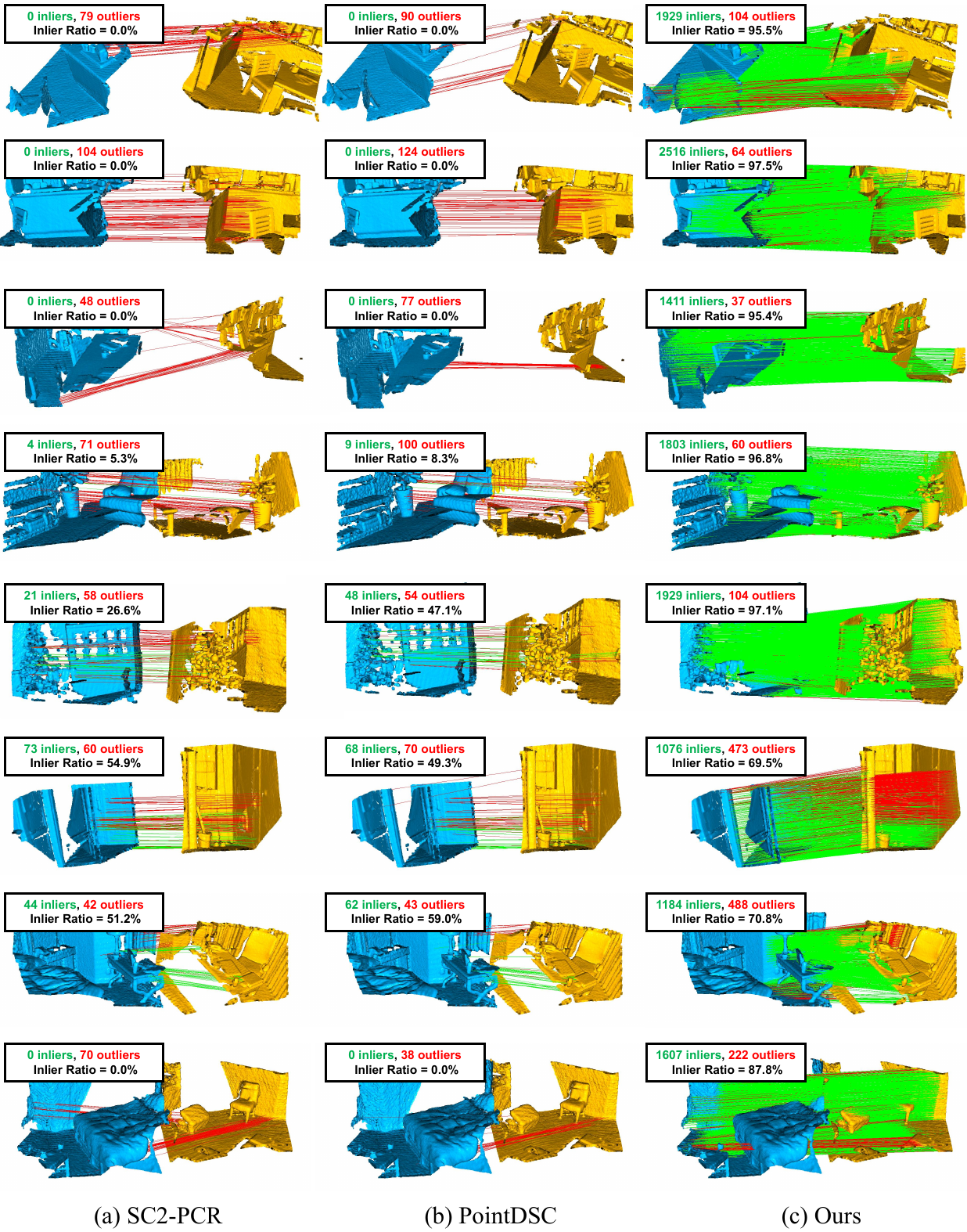}}  
    \caption{
        Correspondence results on 3DMatch.
       }
    \label{qr1}
\end{figure*}

\begin{figure*}[ht]
    \centering{\includegraphics[width=0.95\linewidth]{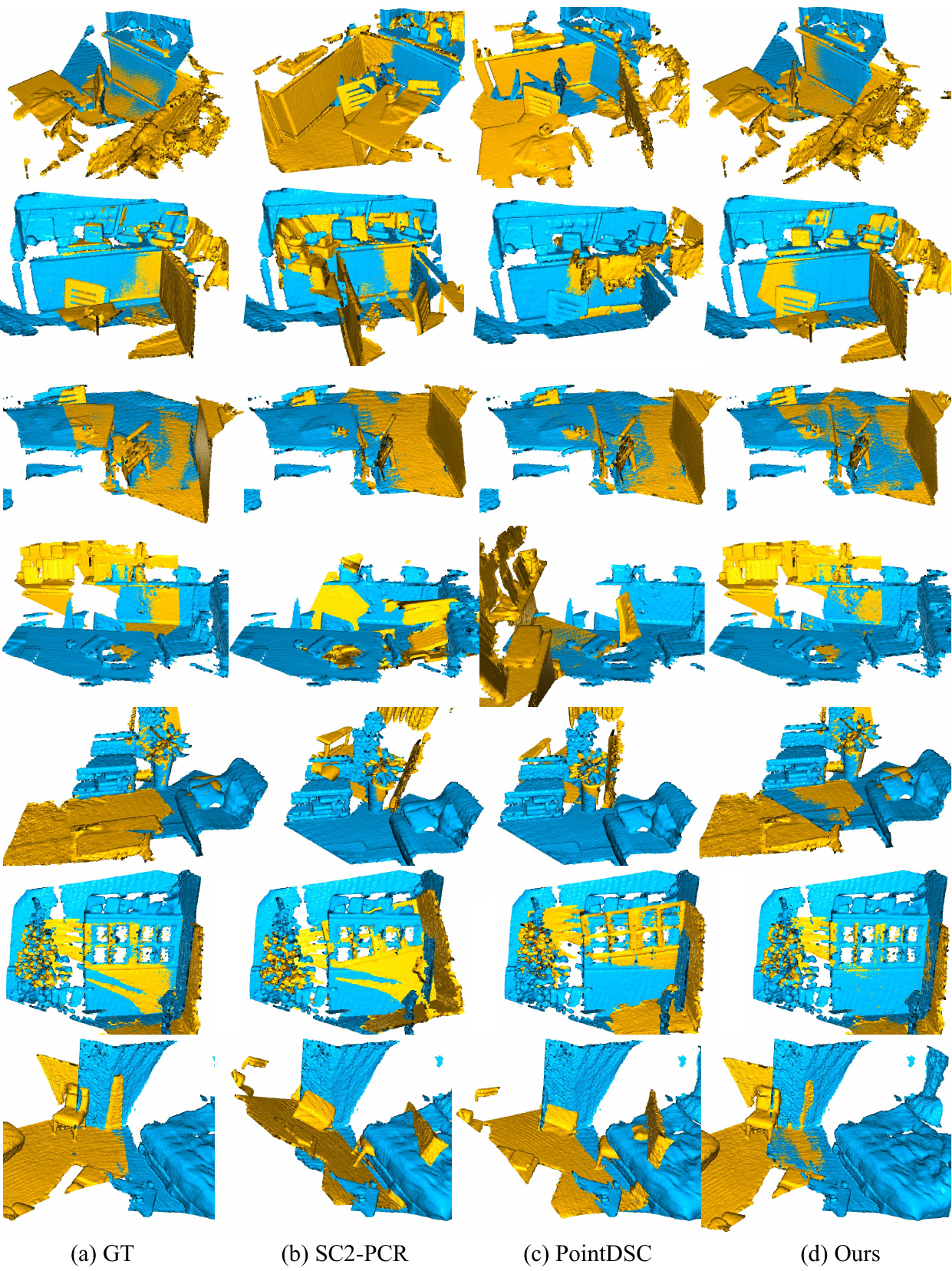}}  
    \caption{
        Registration results on 3DMatch.
       }
    \label{qr2}
\end{figure*}

\ptitle{Additional Qualitative Results on 3DMatch.}
On the 3DMatch dataset, we compare the qualitative results of our method with those of SC2-PCR~\cite{chen2022sc2} and PointDSC~\cite{PointDSC}. The visualization of the correspondences is shown in Figure~\ref{qr1}. As illustrated, when the initial correspondences are of poor quality, outlier removal methods~\cite{fischler1981random,chen2022sc2,PointDSC} struggle to establish high-quality correspondences, which significantly hinders accurate registration. In contrast, our Regor progressively refines correspondences, generating dense and reliable inliers. This provides a robust foundation for accurate 6DoF pose estimation and demonstrates the adaptability of our approach to scenarios with extremely high outlier rates.

Figure~\ref{qr2} presents the qualitative results and registration visualizations. In some scenarios, SC2-PCR~\cite{chen2022sc2} and PointDSC~\cite{PointDSC} fail to achieve robust registration due to the limitations in correspondence quality. Compared to outlier removal methods~\cite{fischler1981random,chen2022sc2,PointDSC}, our method not only achieves robust registration but also yields lower pose errors.

\begin{figure*}[ht]
    \centering{\includegraphics[width=1.0\linewidth]{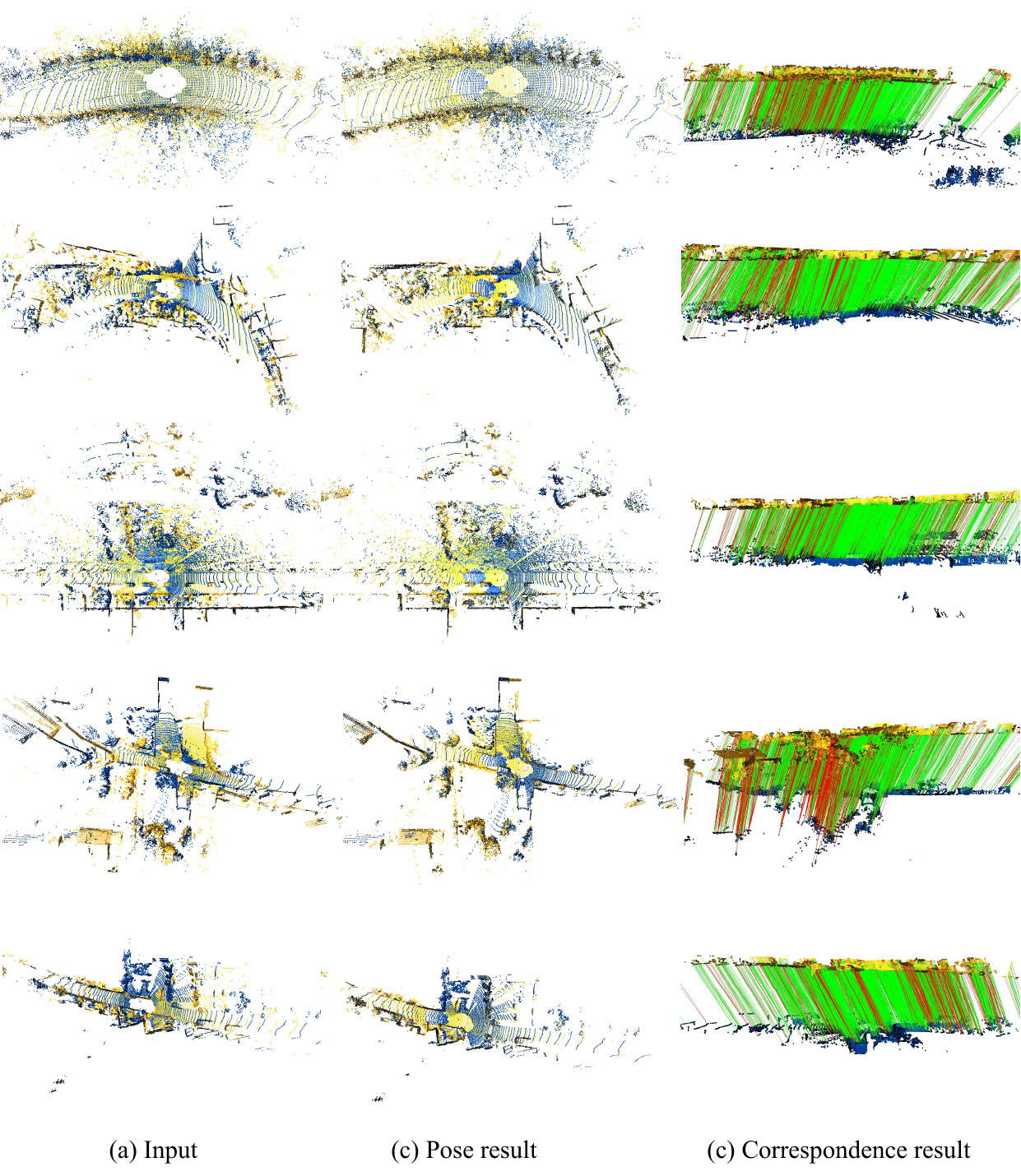}}  
    \caption{
        Qualitative results on KITTI odometry.
       }
    \label{qr3}
\end{figure*}

\ptitle{Additional Qualitative Results on KITTI.}
On the KITTI dataset, we use FPFH~\cite{FPFH} for feature extraction and the results of the qualitative experiments are shown in Figure~\ref{qr3}. Our method achieves accurate and robust registration due to high-quality dense correspondences.


\end{document}